\documentclass[a4paper, 10 pt, onecolumn]{article}
\usepackage[letterpaper,
            left=1in,right=1in,top=1in,bottom=1in,%
            footskip=.25in]{geometry}
            
\pdfoutput=1 

\usepackage[numbers]{natbib}
\usepackage[utf8]{inputenc}
\usepackage{amsthm}
\usepackage{bm}
\usepackage{amsmath, amssymb, url}

\usepackage{fixltx2e}
\usepackage{graphicx}
\usepackage[usenames]{color}

\usepackage{multicol}

\usepackage{graphicx}
\usepackage{amsmath,amsthm,amsfonts,amssymb}
\usepackage{floatrow}
\usepackage{caption}
\usepackage{placeins}
\usepackage{makecell}
\usepackage{subcaption}
\usepackage{multirow}
\usepackage{array}
\usepackage{mathtools}
\usepackage{color}
\usepackage{bbm}
\usepackage{wasysym}
\usepackage[dvipsnames]{xcolor}
\usepackage{tikz}

\usepackage{enumerate}
\usepackage{epstopdf} 
\usepackage{resizegather}

\usetikzlibrary{arrows}

\usepackage{blkarray}

\usepackage{hyperref}
\usepackage[hyphenbreaks]{breakurl}

\usepackage[affil-it]{authblk}

\title{Safety Challenges for Autonomous Vehicles in the Absence of Connectivity}

\date{}

\author{Akhil Shetty, Mengqiao Yu, Alex Kurzhanskiy, Offer Grembek, \\Hamidreza Tavafoghi, and Pravin Varaiya}

\affil{University of California, Berkeley}

\begin{document}

\maketitle

\begin{abstract}
Autonomous vehicles (AVs)  are promoted as a  technology that will create a future with effortless driving  and virtually no traffic accidents. AV companies claim that, when fully developed,  the technology will eliminate 94\% of all accidents that are caused by human error. These AVs will likely avoid the large number of crashes caused by impaired, distracted or reckless drivers. But there remains a significant proportion of crashes for which no driver is directly responsible. In particular, the absence of connectivity of an AV with its neighboring vehicles (V2V) and the infrastructure (I2V) leads to a lack of information that can induce  such crashes. Since  AV designs today do not require such connectivity,  these crashes would persist in the future. Using prototypical examples motivated by the NHTSA pre-crash scenario typology, we show that fully autonomous vehicles cannot guarantee safety in the absence of connectivity. Combining theoretical models and empirical data, we also argue that such hazardous scenarios will occur with a significantly high probability. This suggests that incorporating connectivity is an essential step on the path towards  safe AV technology.
\end{abstract}

\section{Introduction}

\subsection{Background}
With the introduction of autonomous vehicle (AV) technology, the vision of a safe transportation system with effortless driving seemed within reach. The vision  captured the imagination of both venture capitalists and established automobile companies. Within  three years from August 2014 to June 2017 the AV industry attracted more than \$80 billion dollars \cite{Brookings}, so over the decade at least \$100 billion dollars have been invested, exceeding the cost of the Apollo program. 
The AV companies are all pursuing the same goal: a \emph{fully autonomous} vehicle that does not communicate with either infrastructure or neighboring vehicles in order to function safely \cite{sae2014taxonomy}. In 2012, Google co-founder Sergey Brin predicted AVs would be widely available in five years, and AV companies shared the belief that the goal would be realized by 2020 \cite{Cruise2020, Toyota2020, Waymo2020, Brin2012}. 

While limited testing in selected suburban areas and operational design domains (ODDs) has started over the last couple of years, it now seems that the goal of widespread deployment of AVs will be out of reach for the next few decades \cite{kannan2020autonomous}. Over time, the overwhelming complexities involved in urban driving have become apparent. In order to be safe, the AV needs to routinely navigate through out-of-the-ordinary road conditions, unseen obstacles, vehicles with conflicting paths, and inattentive pedestrians. John Krafcik, the CEO of Waymo, recently declared that full autonomy may never be achieved \cite{Krafcik_level5}.  Kyle Vogt, the President and CTO of Cruise, argued that in urban driving an AV must be able to hand off control to a human safety driver for the foreseeable future \cite{cruise_vogtblog}. However, the requirement of a safety driver in urban environments drastically undermines the AV business case.

Currently, the central approach followed by AV companies is to break the complex driving task into sub-tasks of sensing, perception, behavior prediction, planning and control. The expectation is that as these sub-tasks are solved with greater precision, we get closer to eliminating all traffic crashes. In particular, AV companies claim that they can eventually eliminate 94\% of all crashes caused by human error, quoting a report by the National Highway Traffic Safety Administration (NHTSA) \cite{nhtsa2008national} from 2008. However, the report only states the following: \emph{``The critical reason was assigned to drivers in an estimated 2,046,000 crashes that comprise 94 percent of the NMVCCS crashes at the national level. However, in none of these cases was the assignment intended to blame the driver for causing the crash.''} In other words, the 94\% of all crashes attributed to human error involve not only impaired or distracted driving, but also such causes as “false assumption of other’s actions”, “decision error”, “recognition error” and “inadequate surveillance”. Clearly, not all such crashes are simply a consequence of having reckless and error-prone humans at the wheel. This suggests that the actual fraction of crashes that AVs can hope to eliminate is likely to be much lower than 94\%. 

As argued in \cite{koopman_blog}, the fraction of crashes due to overt human error such as impaired driving or violating traffic rules is about 50\%. While AVs make overt errors at present, it is conceivable that they can eliminate such crashes as their sensing and perception technology improves over time.  Indeed, a 50\% reduction in crashes would be a significant safety improvement over the status quo and clearly demonstrates the immense potential that AVs exhibit in improving traffic safety.  However, this still leaves half of all current crashes that are seemingly the result of unavoidable hazardous traffic scenarios which vehicles routinely find themselves in.  
Analyzing an AV's safety capabilities in such hazardous scenarios that do not involve obvious human error is a crucial prerequisite for understanding the potential safety gains they can bring about.

\subsection{Scenario Based Safety Assessment}
A recent NHTSA report \cite{swanson2019statistics} provides a crash typology that describes the diversity of hazardous scenarios commonly observed on the roads. It classifies crashes into 36 scenario types based on the set of events leading up to the crash. For example, the crash type \emph{Backing into Vehicle} includes all crashes in which a vehicle collides with another vehicle while backing. The report also contains the statistical distribution of crashes across these types. This can serve as a useful tool in understanding why crashes occur, how frequently they occur and what type of crashes can be avoided by AVs.

Based on the above NHTSA crash typology, a few crash scenarios are likely to be particularly challenging for AVs. For instance, \emph{Crossing Paths} is a grouping of 6 crash types involving two vehicles moving perpendicular to each other with conflicting paths--a common occurrence at intersections. This group accounts for 19\% of all crashes and damages of $\$135.4$ billion each year. A noteworthy aspect of such crashes is that it is often unclear which vehicle is at fault. There could be static obstacles or surrounding vehicles occluding the field of view of both parties involved so that they do not see each other until it is too late. 
NHTSA's analysis shows that intersection-related crashes are almost 335 times as likely to have ``turned with obstructed view'' as the critical reason than non-intersection-related crashes \cite{nhtsa2010crashfactor}.
\emph{Crossing Paths} crashes could also occur due to one or more vehicles violating traffic rules such that their paths intersect. Even if AVs have perfect sensing and perception capabilities, they will not be immune to occlusions and as a result, will end up in such hazardous scenarios. \emph{Changing Lanes} is another crash grouping that accounts for 12\% of all crashes and damages of $\$32.9$ billion every year. Lane changing involves searching for large enough gaps in traffic in order to safely complete the maneuver. Such large gaps may not always be available, especially during peak hours. In such circumstances with small traffic gaps, the lag vehicle either cooperates or is forced to create the required gap so that the merging vehicle can successfully change lanes. Thus, inaccurate predictions of the lag vehicle behavior can lead to dangerous scenarios which ultimately lead to crashes.

Such a scenario-based assessment of safety is reflected in a recent safety report by Waymo \cite{schwall2020waymo}. It suggests that while AVs rarely make overt errors, they still get into crashes due to the actions and behavior of neighboring vehicles. 
At present, it is unclear whether AVs are capable of navigating safely through such precarious scenarios. The typically reported AV safety statistics such as crashes or disengagements per vehicle mile travelled (VMT) are not very informative as they do not provide sufficient insight on how AVs perform in specific crash scenarios. Moreover, the limited on-road deployment of AVs takes place in ODDs and environments where current AV technology is believed to be ``safe enough''. In such environments, AVs may not encounter many of the above discussed challenging scenarios often enough to gauge their safety performance. For instance, Waymo mainly operates in Phoenix suburbs and only serves 5-10\% of ride requests with its fully driverless fleets \cite{WP-waymo}. The rest of the ride requests are served using fleets with safety drivers. However, hazardous crash scenarios involving occlusions, small traffic gaps and traffic violations are commonplace in dense urban settings. What fraction of such crashes will be eliminated with the introduction of AVs remains an open question.
An alternative technology provides some clues to answer this question. The last couple of decades have seen the rapid adoption of Advanced Driver Assistance Systems (ADAS) that share many features with current AV technologies; for example, Automatic Emergency Braking (AEB), Lane Keeping Assist (LKA), and Adaptive Cruise Control (ACC). Like AV technology, these systems aim to eliminate human error by assisting or warning the driver (e.g., Forward Collision Warning) or by automating driving for a well specified task (e.g., Automatic Parking). Each ADAS addresses a certain type of crash scenario. For example, AEB and LKA reduce rear end and lane departure crashes respectively. It seems that for these specific crash scenarios an AV cannot do significantly better than the corresponding ADAS. Thus, ADAS crash reduction studies could provide a good estimate for potential crash reduction due to AV technology. One such study analyzed the field effectiveness of General Motors ADAS based on safety data from 3.7 million vehicles \cite{UMTRI_GM}. It found a crash reduction rate ranging from 3\% to 81\% depending on the ADAS considered. Notably, while ADAS help improve traffic safety, they fail to prevent a significant proportion of system-relevant crashes. This might presage a similar outcome in the future with AVs. Another study predicting AV-related road traffic fatalities using the German In-Depth Accident Study database arrived at the same conclusion \cite{lubbe2018predicted}.

\subsection{Resolving Information Gaps}
A fundamental prerequisite for an AV to be safe on the roads is an accurate understanding of the positions and behavior of neighboring vehicles, both of which might be partly unknown to it. A common theme among the challenging crash scenarios we discuss above is the presence of certain gaps in the AV's information about these positions and behaviors. Occlusions present a case wherein uncertainty about the location of neighboring vehicles or pedestrians can lead to a crash. Merging is an instance in which not knowing the future behavior of neighboring vehicles (i.e., whether they allow you to merge) is potentially hazardous. 
To tackle such situations, the predominant approach in the AV industry currently is to to arrive at the best possible estimate for the state and behavior of neighboring vehicles given whatever the AV can observe. While this approach may provide satisfactory performance in many cases, it is unlikely that this is a complete solution to address the challenge of information gaps. Indeed, evidence from AV testing supports this observation \cite{av_left_turn_telegraph, av_left_turn_theinfo, verge_krafcik}. 

There are two alternate approaches proposed in the AV literature to overcome such information gaps. Researchers from Mobileye, an AV company acquired by Intel, proposed the Responsibility Sensitive Safety (RSS) framework to specifically deal with this challenge \cite{shalev2017formal}. The key idea is to limit an AV's maneuvers such that it is robust to all reasonable states and behaviors that are consistent with its partial observations, thus ensuring safety. However, guaranteeing safety alone is not desirable if it comes at the cost of a sharp drop in throughput. An extreme example that illustrates this point is the recommendation for an AV to not move out of the garage -- this is guaranteed to be safe, but is clearly untenable. It is not immediately clear whether the RSS framework can guarantee safety in the presence of information gaps without a substantial reduction in throughput.

Another approach to deal with information gaps is to explicitly bridge them using either infrastructure-to-vehicle (I2V) or vehicle-to-vehicle (V2V) connectivity. In the case of occlusions and traffic violations at intersections, I2V communication would  ensure that the involved vehicles detect each other in time \cite{grembek2019making, misener2010cooperative, sae-j2735}. For the lane changing case, V2V communication would ensure that the involved parties are in agreement, thus preventing a potential hazardous situation \cite{Hsu1991,luo2016dynamic, yang2004vehicle}. 
However, reliance on connectivity is costly. Forgoing connectivity allows companies to deploy their vehicles on the roads without having to wait for the required sensors and communication channels being set up. Moreover, security concerns arise with dependency on surrounding vehicles or infrastructure for information \cite{cui2019review, mccluskey2017connected}. Thus, the question arises whether it is possible to avoid crashes in these scenarios without relying on connectivity. 

\subsection{Our Contribution}
In this paper, we develop a framework to study hazardous scenarios involving information gaps that routinely occur in dense urban environments. In particular, we study three groups of scenarios with information gaps: (i) occlusions, (ii) traffic violations, and (iii) behavior prediction uncertainty. 
We argue that in these scenarios, even with perfect sensing and perception capabilities, AVs would still face significant challenges if they want to guarantee a safety level comparable to human baselines. That is, to ensure such a safety level, AVs must significantly limit their space of maneuvers and sacrifice on traffic efficiency and throughput. As such, we also conclude that the worst-case approach, as laid out by Mobileye's RSS framework, does not provide a practical solution to address the safety consideration in these scenarios. Our analysis suggests that a significant portion of crashes due to information gaps would still persist under the current approach followed by AVs.  
Accordingly, we discuss how connectivity with infrastructure or surrounding vehicles can alleviate these safety concerns by directly addressing the information gap in these scenarios.  


\section{Occlusions} 
\label{sec:occlusions}

\begin{figure}[ht]
    \centering
    \includegraphics[scale=0.3]{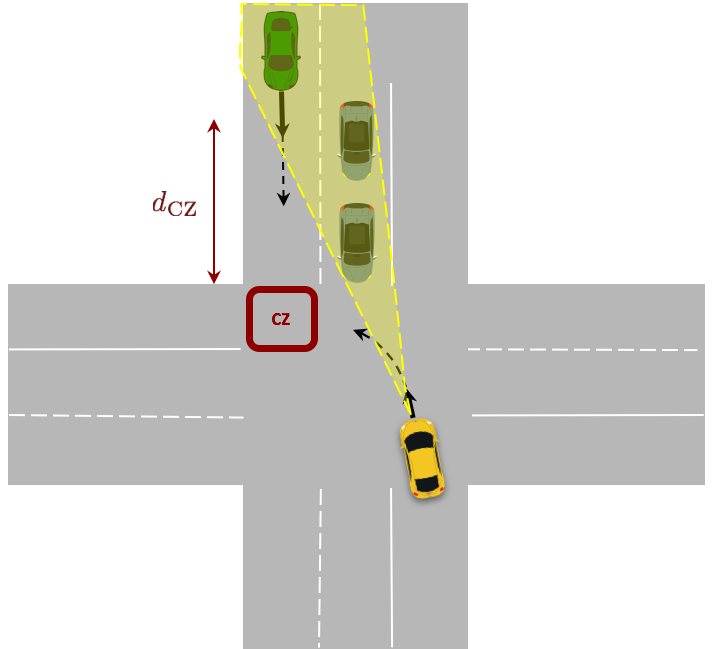}
    \caption{AV (yellow) making unprotected left turn with through moving vehicles (green) occluded by queued up left-turning vehicles (blue). The AV's occluded field of view is illustrated by the shaded yellow region. The red box \emph{CZ} denotes the conflict zone.}
    \label{fig:veh_veh_occlusion}
\end{figure} 

Let us consider a commonly occurring on-road occlusion scenario and analyze how an AV should act in order to be safe. Consider an AV at a signalized traffic intersection on the left-turning lane as illustrated in Figure \ref{fig:veh_veh_occlusion}. We assume that the AV is equipped with sensors (e.g., LIDAR) that provide a 360 degree view of its surroundings. There is no protected left-turn phase, so it is waiting for gaps in the on-coming traffic to make a left turn. The left-turn opposing lane is queued up, thus, blocking the AV's view of through-moving vehicles (TMVs) in the opposing lane.\footnote{As the LIDAR is placed on top of the AV, such occluding vehicles may not always be a problem. A LIDAR will not be able to detect the occluded TMVs and pedestrians if the height at which the LIDAR is placed is lower than the height of nearby occluding vehicles. Suppose the AV is a sedan (average height ~56 inches) with a LIDAR placed one foot above the roof. This would imply that the AV cannot view beyond any vehicle taller than 68 inches, thus resulting in a potential occlusion. This includes light pickup trucks (average height - 76 inches) as well as heavy vehicles which tend to be even taller. Note that light pickup trucks alone account for $\sim$ 20\% of US vehicle market share, thus, such occlusion scenarios are likely to occur frequently.} 

The task of the AV is to choose the right time to make a left-turn so that it does not collide with through-moving vehicles. This scenario falls within \emph{Crash Type 30: Left Turn Across Path/Other Direction (LTAP/OD)} of the NHTSA crash typology and accounts for 5.8\% of all crashes. Furthermore, such crossing path crashes have the highest comprehensive costs and equivalent lives lost among all crash groups in the NHTSA report \cite{swanson2019statistics}. Recognizing the safety costs of such crashes, fleet operators such as UPS design their routes so that they do not involve left turns \cite{holland2017ups}.

We assume that traffic in the opposing through lane is free-flowing and the arrival of TMVs is modeled as a Poisson process with rate $\lambda$. We assume that the through-moving  vehicle can make an evasive maneuver (for e.g., hit the brakes) only after it sees the left-turning AV, else it maintains its current speed. As AVs will have to drive among human drivers when they are introduced on the roads, we assume that all other vehicles are human driven with a reaction time $\rho \in [0.7,2.5]$ \footnote{We use the term \emph{reaction time} to refer to the total time required for perception (mental processing time for recognizing need for evasive action), driver response (time taken to make the evasive maneuver, for e.g., hitting the brakes), and device response (time between driver's action and corresponding vehicle response). This is also referred to as \emph{stopping time} in the literature \cite{green2000long}.}. We maintain this assumption for the rest of the paper. We use the following values for relevant intersection geometry and vehicle parameters:
\begin{itemize}
\item Lane width - 4 m,
\item Vehicle length - 4 m,
\item Vehicle width - 2 m,
\item Maximum acceleration rate - 3 m/$\mathrm{s}^2$,
\item Maximum deceleration rate - 4 m/$\mathrm{s}^2$.

\end{itemize}

\subsection{Can the AV make an unprotected left-turn with guaranteed safety?}
We first consider under what conditions an AV can make an unprotected left-turn irrespective of the arrival process of TMVs. In this case, the AV's safety depends on whether the TMV can brake in time to avoid the AV\footnote{Please refer to Appendix A for a detailed analysis of why an AV cannot ensure its safety in this scenario even if it attempts to make an evasive maneuver.}. The worst  case for the AV in such a situation is when the TMV is just beyond the AV's field of view when it decides to make a left-turn. We use the term \emph{conflict zone} to refer to the region of the intersection where the paths of the left-turning and through moving vehicles intersect -- marked in red in Figure \ref{fig:veh_veh_occlusion}. Let $v_{\mathrm{th}}$ and $a_{\mathrm{dec}}$ denote the velocity and maximum deceleration rate of the TMVs. Let $d_{\mathrm{CZ}}$ denote the distance from conflict zone at which the AV comes into the field of view of the TMV. Based on the above parameters for intersection geometry and vehicle dimensions, $d_{\mathrm{CZ}}$ turns out to be 12 m. As the TMV takes time $\rho$ to react to the left-turning AV, its distance from the conflict zone when it starts decelerating is $d_{\mathrm{CZ}} - v_{\mathrm{th}}\rho$. Assuming the TMV brakes at the maximum rate $a_{\mathrm{dec}}$, it can avoid a crash with the AV so long as 
\begin{align} \label{eq:max_TMV_velocity}
    v_{\mathrm{th}}^2 \leq 2 a_{\mathrm{dec}} (d_{\mathrm{CZ}} - v_{\mathrm{th}}\rho).
\end{align}

\begin{figure*}[h!]
  \centering
  \begin{subfigure}[b]{0.34\linewidth} 
    \includegraphics[width=\linewidth]{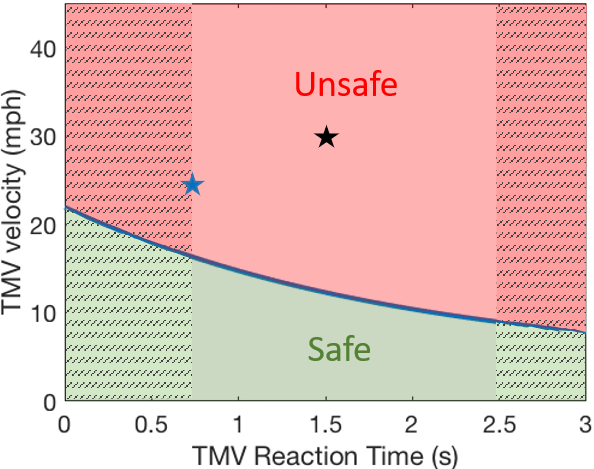}
    \label{fig:lambda_4cases}
  \end{subfigure} \hspace{0.7cm}
  \begin{subfigure}[b]{0.33\linewidth} 
    \includegraphics[width=\linewidth]{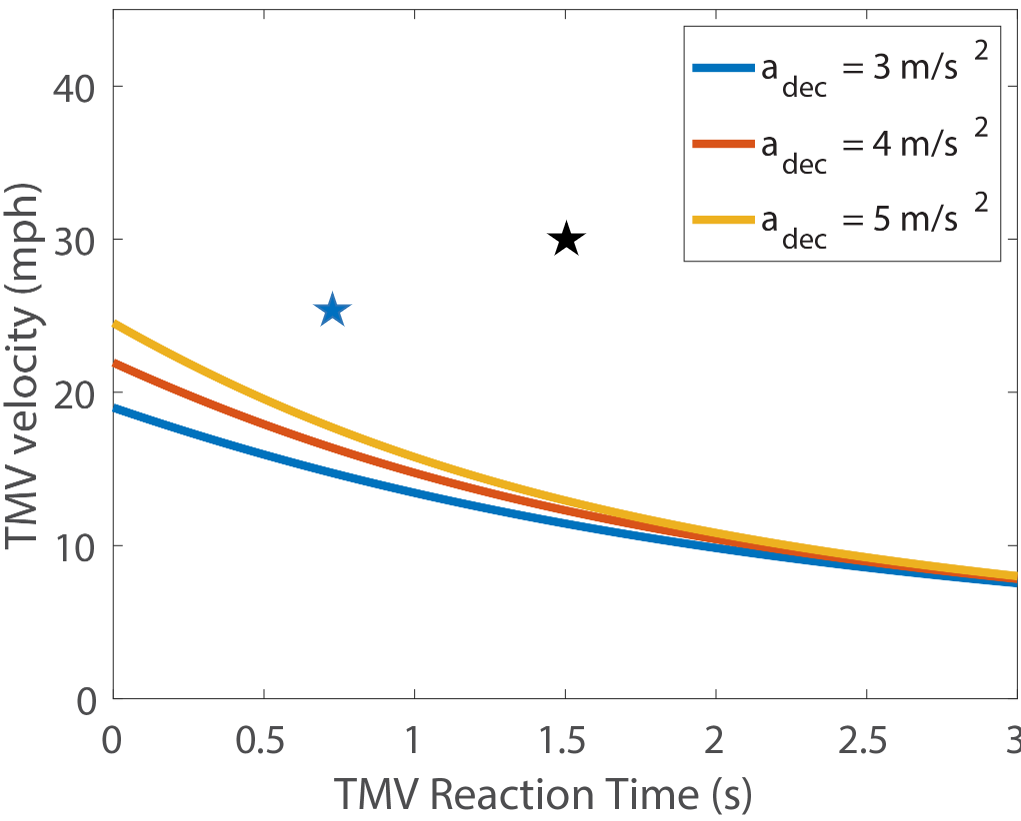}
    \label{fig:N_4cases}
  \end{subfigure} 
  \caption{Safe TMV velocity as a function of TMV reaction time and deceleration rate. \\ 
 (Left:) Pairs of ($v_{\mathrm{th}}, \rho$) that ensures safety are in green, whereas the unsafe pairs are in red. Typical human reaction times varies between 0.7\:s and 2.5\:s (unshaded region). The blue star (\textcolor{blue}{$\star$}) denotes the optimistic ($v_{\mathrm{th}} = 25 \ \mathrm{mph}, \rho = 0.7$ s) pair considered in our analysis, while the black star ($\star$) denotes typical observed values ($v_{\mathrm{th}} = 30\:\mathrm{mph}, \rho = 1.5$\:s) in practice. Note that both points are in the unsafe region. \\
 (Right:) Change in the safe region boundary (as shown in left) for different values of deceleration rate $a_{\mathrm{dec}}$. The higher the deceleration rate, the higher is the maximum TMV velocity that can still ensure safety. Note that even for higher values of deceleration rate, both the optimistic and typical values are in the unsafe region.}
 \label{fig:occlusion_sensitivity}
\end{figure*}

Even under the most optimistic choice for TMV reaction time $\rho = 0.7 \ s$ \cite{green2000long}, an AV can be guaranteed to be safe only if the TMVs are moving at no more than $17$ mph. This is much lower than typical through moving vehicle speeds observed at intersections, especially because they have right-of-way in such a setting. Additionally, Figure \ref{fig:occlusion_sensitivity} shows the sensitivity of our analysis with respect to different choices of reaction time $\rho$ and deceleration rate $a_{\mathrm{dec}}$. This suggests that an AV cannot guarantee safety while making an unprotected left-turn under such occlusions.

\subsection{Can the AV be ``safe enough" while making an unprotected left-turn?} \label{subsec:veh_veh_occlusion}
Realizing that it is impossible to guarantee safety in this setting, we consider an AV that is willing to accept a probability of collision no greater than $p_{\mathrm{coll}}$ under such an occlusion. Note that the arrival rate $\lambda$ of TMVs is initially unknown to the AV. Let us assume a through-movement velocity $v_{\mathrm{th}} = 25$ mph. Recall that $d_{\mathrm{CZ}}$ is the TMV's distance from the conflict zone when the AV comes into its field of view. Let $d^{\mathrm{min}}_{\mathrm{CZ}}$ denote the minimum distance such that the TMV can brake to a stop before reaching the conflict zone. If $d_{\mathrm{CZ}} < d^{\mathrm{min}}_{\mathrm{CZ}}$, then we have a potential collision. Although we account for the TMVs attempting to brake when they see the AV, we do not consider more complicated evasive maneuvers like swerving that could also be used by the TMVs. Such scenarios in which a collision would occur unless evasive maneuvers are carried out by the involved vehicles are called \emph{traffic conflicts}.\footnote{The \emph{traffic conflicts} technique was introduced by Perkins and Harris as a surrogate measure of traffic safety \cite{perkins1968traffic}. A traffic conflict is defined as an event that would lead to a crash unless an evasive action such as braking or swerving is taken. As traffic crashes are very rare events, actual crash data is scarce and unreliable as a safety metric. On the other hand, traffic conflicts are abundant and are amenable to empirical estimation. The estimated number of traffic conflicts can then be multiplied by a suitable factor depending on the traffic scenario to get the predicted number of traffic crashes.} It is reasonable to expect that at least some of these traffic conflicts will result in crashes. We allow for the possibility of other evasive maneuvers being used by the TMVs by introducing $\gamma$—the ratio of traffic conflicts to collisions. We analyze the probability of a traffic conflict and then translate it to a collision probability using the following equation:
\begin{align}
    p_{\mathrm{conf}} = \gamma p_{\mathrm{coll}}.\label{eq:conflict_to_collision}
\end{align}
Note that the conflict-to-collision ratio $\gamma$ depends on the scenario considered. For our analysis, we use $\gamma = 1490$ based on an empirical estimate of opposing left-turn conflict-to-collision ratio in \cite{glauz1985expected}.\footnote{Table 8 in \cite{glauz1985expected} contains conflict-to-collision ratios for traffic scenarios such as opposing left-turn, left-turn same direction and through cross traffic under varying traffic volumes. Note that the conflict-to-collision ratio $\gamma$ is the inverse of the accident/conflict ratio in Table 8.} Given that the TMV arrivals are Poisson \footnote{Modeling the traffic flow as a Poisson process is standard in the literature \cite[Ch. 9]{TFT}, \cite{TRB}. In our analysis we assume that the arrival rate of vehicles are constant and does not change over time. As such, we simplify the estimation problem AVs face to a single scalar parameter estimation. We note that one can consider a more complex model to captures the dynamically varying flow rate at an intersection. However, considering such a more complex model requires the estimation of multiple parameters, and thus, more observation time for AV to estimate the model.}, the probability of traffic conflict can be translated into the probability that there is a Poisson arrival in an interval of $t_{\mathrm{conf}}$ sec, where
\begin{align}
  t_{\mathrm{conf}} = \frac{d^{\mathrm{min}}_{\mathrm{CZ}} - d_{\mathrm{CZ}}}{v_{\mathrm{th}}}.
\end{align}
If the AV knows $\lambda$, its decision is straightforward. It should make the left-turn if the probability of at least one TMV arrival in a time interval of length $t_{\mathrm{conf}}$ is less than the AV's maximum allowed probability of traffic conflict, i.e., 
\begin{align}
    1 - e^{-\lambda t_{\mathrm{conf}}} \leq p_{\mathrm{conf}}.
\end{align}
Thus, the AV would make the turn if $\lambda \leq \lambda_{\mathrm{max}}$, where
\begin{align}
    \lambda_{\mathrm{max}} = \frac{1}{t_{\mathrm{conf}}}\log\Bigg(\frac{1}{1-p_{\mathrm{conf}}}\Bigg).
\end{align}
However, the AV does not know $\lambda$, which raises the question: \emph{In the time the AV waits at the intersection, can it estimate $\lambda$ with reasonable confidence so that it can make the left-turn when the situation is indeed safe?}
\\ \\
This question can be posed as a hypothesis testing problem:
\begin{align}
    H_0: \lambda \geq \lambda_{\mathrm{max}}, \qquad 
    H_1: \lambda < \lambda_{\mathrm{max}}.
\end{align}
The AV observes through-moving traffic while it waits for $t_{\mathrm{obs}}$ seconds at the intersection. It decides to make the left turn if it can reject $H_0$. For a level-$\alpha$ test, this implies that the AV makes the turn if it sees no TMV arrivals in $t_{\mathrm{obs}}$ seconds, i.e., 
\begin{align}
    e^{-\lambda_{\mathrm{max}} t_{\mathrm{obs}}} \leq \alpha.
\end{align}
Thus, we have
\begin{align}
    t_{\mathrm{obs}} = \frac{1}{\lambda_{\mathrm{max}}}\log \Bigg(\frac{1}{\alpha}\Bigg). \label{eq:t_obs}
\end{align}
It is not evident what an acceptable collision probability $p_{\mathrm{coll}}$ should be. We derive an upper bound $\bar{p}_{\mathrm{coll}}$ for this quantity by choosing the intersection with the highest rate of opposing left-turn crashes in San Francisco as our baseline -- Market Street and Octavia Street with 10 crashes in the period 2011-17 \cite{tims}. Then, 
\begin{align}
    \bar{p}_{\mathrm{coll}} &= \frac{\text{Number of opposing left-turn crashes per year}}{\text{Number of left-turns under occlusion per year}}. 
\end{align}
Assuming that the occlusion scenario occurs during peak hours, we have 
\begin{align}
    \text{Number of left-turns under occlusion per year} &= \text{Left-turn rate} \times \text{Number of peak hours per day}  \\ & \quad \times \text{Number of weekdays in a year} \nonumber.
\end{align}
Assuming a flow of 1000 vehicles/hr through the intersection with 10\% of vehicles turning left on average, gives a left-turn rate of 100 turns/hr. Assuming 4 peak hours/weekday, we get $\bar{p}_{\mathrm{coll}} = 1.4 \times 10^{-5}$. We use this upper bound in our calculations to get a lower bound for $t_{\mathrm{obs}}$ using \eqref{eq:t_obs}. We use $\gamma = 1490$ based on \cite{glauz1985expected}. As $p_{\mathrm{conf}} = \gamma \bar{p}_{\mathrm{coll}} = 2.1 \times 10^{-2}$, we choose $\alpha = 10^{-4}$ for the hypothesis testing problem. As $v_{\mathrm{th}} = 25$ mph (11.18 m/s), we get $d^{\mathrm{min}}_{\mathrm{CZ}} = 23.45 $ m using \eqref{eq:max_TMV_velocity}. From \eqref{eq:t_obs}, we see that the AV needs to observe traffic for $t_{obs} =  443$ seconds in order to make its decision to turn. This is prohibitively high. As a result, we conclude that the AV cannot safely make the turn with the required collision probability without drastically reducing throughput, i.e., waiting at the intersection over multiple traffic cycles until it is confident enough to make the left turn.\footnote{As we already account for braking action by TMVs, the conflict-to-collision ratio $\gamma$ chosen for our analysis is considerably larger than the actual $\gamma$ for the given setting. Thus, the required waiting time at the intersection is even larger than what we have calculated.}

\subsection{Occluded Pedestrians} \label{subsec:occluded_pedestrians}
 The scenario considered in the previous subsections gets  more challenging when pedestrians are involved. Crashes involving pedestrians have the highest fatality rate among all crash groupings in the NHTSA typology \cite{swanson2019statistics}. Such crashes account for 58\% of all fatal crashes near intersections in the city of San Francisco \cite{tims}. In what follows, we investigate the role of occlusions in causing such crashes. 

\begin{figure}[!ht]
  \includegraphics[width=1.0\textwidth]{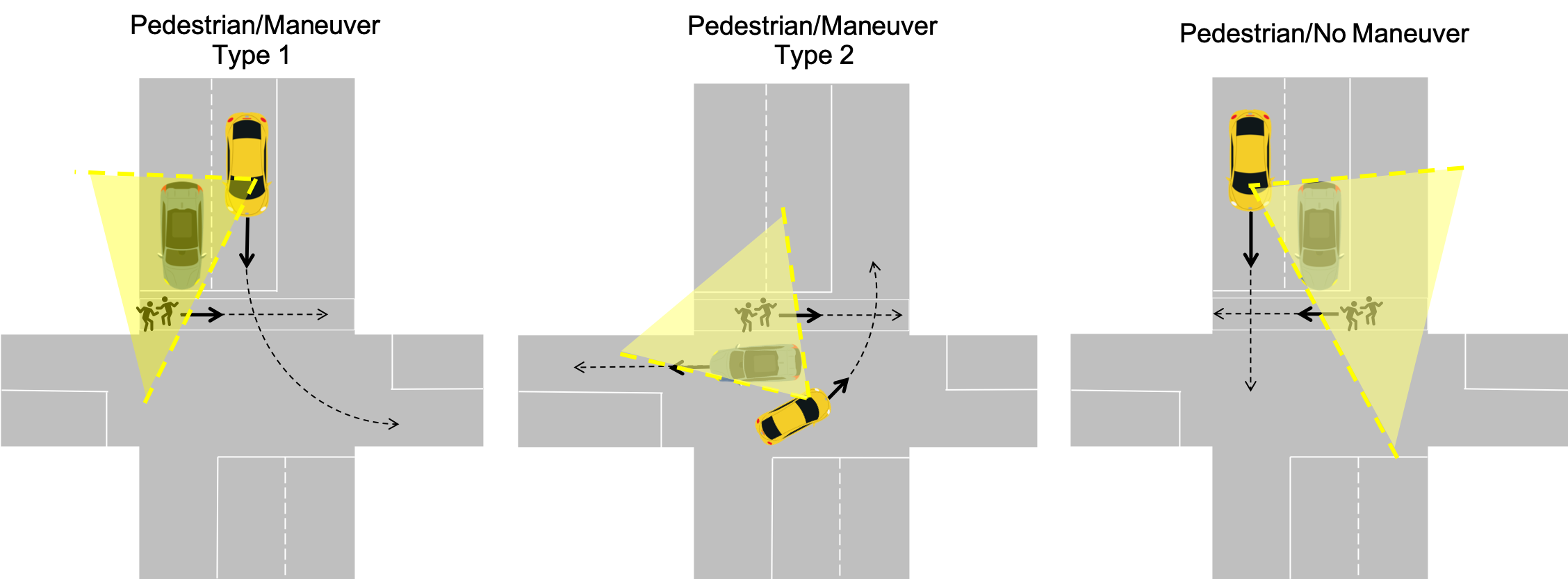}
  \caption{Three pre-crash scenarios with occluded pedestrians: AV (in yellow) is occluded by stopped or moving vehicles (in blue), and the AV's occluded field of view is illustrated by the shaded yellow region.}\label{fig:veh_ped_collisions}
\end{figure}

The NHTSA pre-crash scenario typology includes two crash types to classify vehicle-pedestrian crashes based on whether the vehicle is engaged in a maneuver (e.g., passing, turning): \emph{Crash Type 10: Pedestrian/Maneuver} and \emph{Crash Type 11: Pedestrian/No Maneuver} \cite{swanson2019statistics}. However, it does not specify if occlusions contributed to the crash. Through a further classification of the above crash types based on possible occlusions, we arrive at three pre-crash scenarios in which pedestrians being occluded by a neighboring vehicle leads to a crash as illustrated in Figure \ref{fig:veh_ped_collisions}. 

\begin{itemize}
    \item \textbf{Pedestrian/Maneuver Type 1}: The AV is turning left at the beginning of its green phase while pedestrians are trying to cross as soon as possible since the pedestrian phase is changing from the flashing phase to red. The field of view of the left-turning AV is occluded by queued up vehicles on the adjacent right lane. Such a scenario can occur in both protected and permissive left turn signal phases.
    \item \textbf{Pedestrian/Maneuver Type 2}: As the left-turning AV enters the intersection, it cannot detect pedestrians on the crosswalk as its field of view is obstructed by the through moving vehicles. This results in a crash if the AV cannot brake in time to avoid the pedestrians. Such a situation can arise during the permissive left-turn phase. 
    \item \textbf{Pedestrian/No Maneuver}: The field of view of the through-moving AV is occluded by stopped vehicles on adjacent lanes. At the beginning of the green through phase, the AV might collide with pedestrians who could not finish crossing during the previous phase, especially those who entered the crosswalk during the yellow phase.
    
\end{itemize}

We model pedestrians as Poisson arrivals on the crosswalk moving at a fixed velocity. We assume that they are inattentive and as a result, do not attempt to evade the AV. Although this is not always the case, such inattentive behavior is commonly observed during changes in signal phase when pedestrians hastily attempt to cross the road. We use the term \emph{pedestrian conflict zone} to refer to the region of crosswalk intersecting with the AV's path. For our analysis, $t=0$ is the time at which the AV first detects the pedestrian. We introduce the following notation:
\begin{itemize}
    \item $\lambda_{\mathrm{ped}}$: arrival rate (ped/s) of pedestrians,
    \item $v_{\mathrm{ped}}$: pedestrian speed (m/s) when crossing on yellow/flashing phase/at the end of pedestrian phase,
    \item $D_{\mathrm{ped\_to\_crash}}$: distance (m) from pedestrian's position at $t=0$ to the center of pedestrian conflict zone,
    \item $v_{\mathrm{AV}}$: speed of the AV (m/s),
    \item $D_{\mathrm{veh\_to\_crash}}$: distance (m) from vehicle's position at $t=0$ to the pedestrian conflict zone,
    \item $a_{\mathrm{acc}}$: maximum AV acceleration rate ($\mathrm{m/s^2}$),
    \item $a_{\mathrm{dec}}$: maximum AV deceleration rate ($\mathrm{m/s^2}$),
    \item $w_{\mathrm{AV}}$: width of AV (m).
\end{itemize}
We begin our analysis with the \emph{Pedestrian/Maneuver Type 2} scenario, and then show how this approach can be suitably modified to accommodate the other two scenarios. Notice that the following conditions need to be satisfied for a \emph{Pedestrian/Maneuver Type 2} scenario to end up in a crash:
\begin{itemize}
    \item \textbf{Condition 1:} The AV's view of the crosswalk is obstructed by passing through-moving vehicles when it is making an unprotected left-turn. 
    \item \textbf{Condition 2:} When the AV finally detects the pedestrian, it does not have enough time to brake to a stop before the conflict zone.
\end{itemize}
\begin{figure}[ht] 
    \centering
    \includegraphics[scale=0.3]{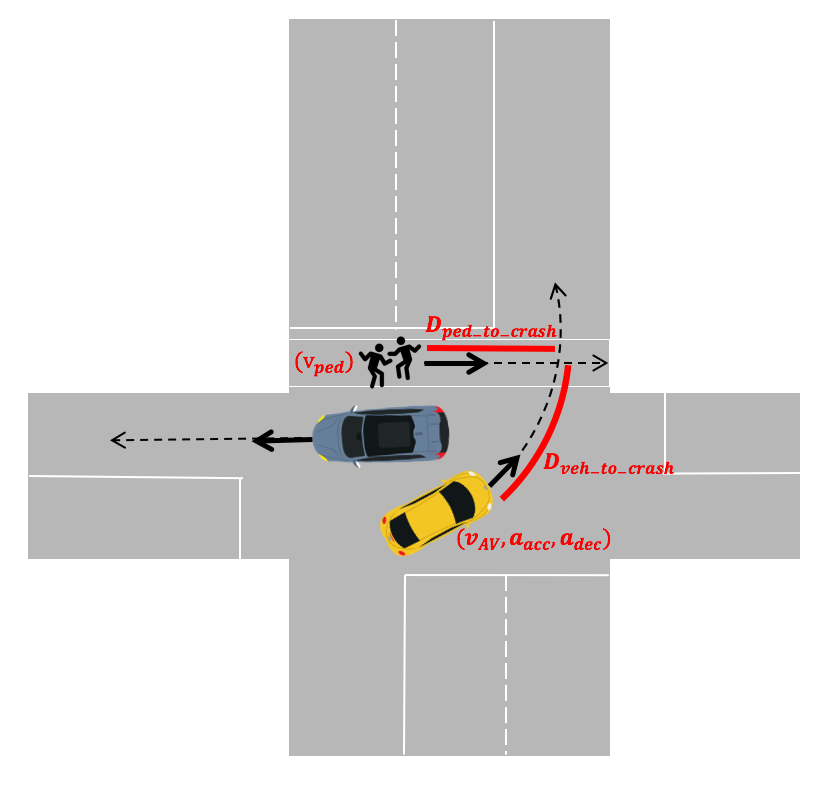}
    \caption{Calculating conflict probability for Pedestrian/Maneuver Type 2.}
    \label{fig:ped_scenario2}
\end{figure}
Assuming that both these conditions hold, a crash can occur if the pedestrian's original trajectory intersects that of the AV. When the AV first detects the pedestrian, it has two choices depending on the location of the pedestrian on the crosswalk: decelerate so that the pedestrian can cross the conflict zone before it reaches there or accelerate and cross the conflict zone before the pedestrian arrives there. As we do not account for pedestrians being attentive or more complicated maneuvers such as swerving by the AV, we classify such a scenario as a traffic conflict, as in Section \ref{subsec:veh_veh_occlusion}. Let $\delta$ denote the time required by pedestrian to cross the conflict zone, i.e., $\delta = w_{\mathrm{AV}}/v_{\mathrm{ped}}$. Let $t_{\mathrm{acc}}$ denote the time taken by the AV to reach the conflict zone while it accelerates. Then,
\begin{equation*}
   D_{\mathrm{veh\_to\_crash}} =  v_{\mathrm{AV}}t_{\mathrm{acc}} + \frac{1}{2}
   {a_{\mathrm{acc}}} t_{\mathrm{acc}} ^2.
\end{equation*}
Taking the positive root of this quadratic equation gives
\begin{align}
    t_{\mathrm{acc}} = \frac{\sqrt{2a_{\mathrm{acc}}D_{\mathrm{veh\_to\_crash}} + v_{\mathrm{AV}}^2}-v_{\mathrm{AV}}}{a_{\mathrm{acc}}},
    \label{eq:t_acc}
\end{align}
and a conflict occurs if the pedestrian arrives at the center of the conflict zone in the time interval $ \mathcal{T}_{\mathrm{acc}} = [t_{\mathrm{acc}} - \delta/2, t_{\mathrm{acc}} + \delta/2]$. 

On the other hand, let $t_{\mathrm{dec}}$ denote the time taken by the AV to reach the conflict zone while it decelerates. Then, instead of (\ref{eq:t_acc}), we have
\begin{align}
    t_{\mathrm{dec}} = \frac{v_{\mathrm{AV}}-\sqrt{v_{\mathrm{AV}}^2 -2a_{\mathrm{dec}}D_{\mathrm{veh\_to\_crash}}}}{a_{\mathrm{dec}}},
    \label{eq:t_dec}
\end{align}
and a conflict occurs if the pedestrian arrives at the center of the pedestrian conflict zone in the time interval $\mathcal{T}_{\mathrm{dec}} = [t_{\mathrm{dec}} - \delta/2, t_{\mathrm{dec}} + \delta/2]$. Thus, a conflict is unavoidable if the pedestrian arrives in the time interval $\mathcal{T}_{\mathrm{acc}} \cap \mathcal{T}_{\mathrm{dec}} = [t_{\mathrm{dec}} - \delta/2, t_{\mathrm{acc}} + \delta/2]$. This translates into the following condition on $D_{\mathrm{ped\_to\_crash}}$ \footnote{Please refer to Appendix B for a more detailed explanation.}: 
\begin{align}
    D_{\mathrm{ped\_to\_crash}} \in [(t_{\mathrm{dec}}-\delta/2)  v_{\mathrm{ped}}), \  (t_{\mathrm{acc}}+\delta/2)v_{\mathrm{ped}}].
    \label{eq:d_ped_to_crash}
\end{align}
The probability of this event can be interpreted as the probability of arrival of at least one pedestrian in a time interval of length $(t_{\mathrm{acc}} - t_{\mathrm{dec}} + \delta)$. As the pedestrian arrival process is Poisson with rate $\lambda_{\mathrm{ped}}$, we have 
\begin{align}
    P_{\text{scenario2}}(\text{Conflict $|$ Conditions 1 and 2 hold})
    &=1 - e^{-\lambda_{\mathrm{ped}}(t_{\mathrm{acc}} - t_{\mathrm{dec}} + \delta)}.
    \label{eq:p_scenario2}
\end{align}
Based on intersection geometry and commonly observed values for model parameters, we set these values: $v_{\mathrm{ped}}=2$ m/s, $\lambda_{\mathrm{ped}}=\frac{1}{60}$ ped/s, $w_{\mathrm{AV}} = 2$ m, $v_{\mathrm{AV}}= 15 $ mph (6.71 m/s), $D_{\mathrm{veh\_to\_crash}}=4$ m, $a_{\mathrm{acc}}=3 \mathrm{ \ m/s^2}$, \text{ and } $a_{\mathrm{dec}}=4 \mathrm{ \ m/s^2}$. Then,  $\delta = w_{\mathrm{AV}}/v_{\mathrm{ped}} = 1$ sec. Using (\ref{eq:t_acc}), (\ref{eq:t_dec}), (\ref{eq:d_ped_to_crash}), and (\ref{eq:p_scenario2}), we get
\begin{align}
    &D_{\mathrm{ped\_to\_crash}} \in [0.55, 2.07]\ \mathrm{m}, \nonumber\\
    &P_{\text{scenario2}}(\text{Conflict $|$ Conditions 1 and 2 hold})=0.0125. 
\end{align}
The above analysis can be conveniently modified for the other two scenarios. For the \emph{Pedestrian/No Maneuver} scenario, Condition 1 should be rephrased as: \emph{The AV's view of the crosswalk is occluded by stopped vehicles in the adjacent lane when it is passing through the intersection at the beginning of the green phase}. Condition 2 remains the same. As in the \emph{Pedestrian/Maneuver Type 2} scenario, the AV can either accelerate or decelerate to avert a crash. Notice that the expressions \eqref{eq:t_acc} and \eqref{eq:t_dec} for $t_{\mathrm{acc}}$ and $t_{\mathrm{dec}}$ respectively remain the same for this scenario. The main difference is in the relevant intersection and vehicle parameters. As the AV is through moving and has right of way, we have $v_{\mathrm{AV}} = 25$ mph (11.18 m/s) and $D_{\mathrm{veh\_to\_crash}}=4$ m. The rest of the parameters remain the same as above. 
Using (\ref{eq:t_acc}), (\ref{eq:t_dec}),(\ref{eq:d_ped_to_crash}), and (\ref{eq:p_scenario2}), gives
\begin{align}
    &D_{\mathrm{ped\_to\_crash}} \in [0, 1.68] \ \mathrm{m}, \nonumber\\
    &P_{\text{scenario3}}(\text{Conflict $|$ Conditions 1 and 2 hold})=0.0158. 
\end{align}
Similarly, for \emph{Pedestrian/Maneuver Type 1}, we set $v_{\mathrm{AV}}=6.71$ m/s and $D_{\mathrm{veh\_to\_crash}}= 3$ m. Then, we have
\begin{align}
    &D_{\mathrm{ped\_to\_crash}} \in [0.063, 1.82] \ \mathrm{m}, \nonumber\\
    &P_{\text{scenario1}}(\text{Conflict $|$ Conditions 1 and 2 hold})=0.0145. 
\end{align}

\begin{figure*}[h!]
  \centering
  \begin{subfigure}[b]{0.4\linewidth} 
    \includegraphics[width=\linewidth]{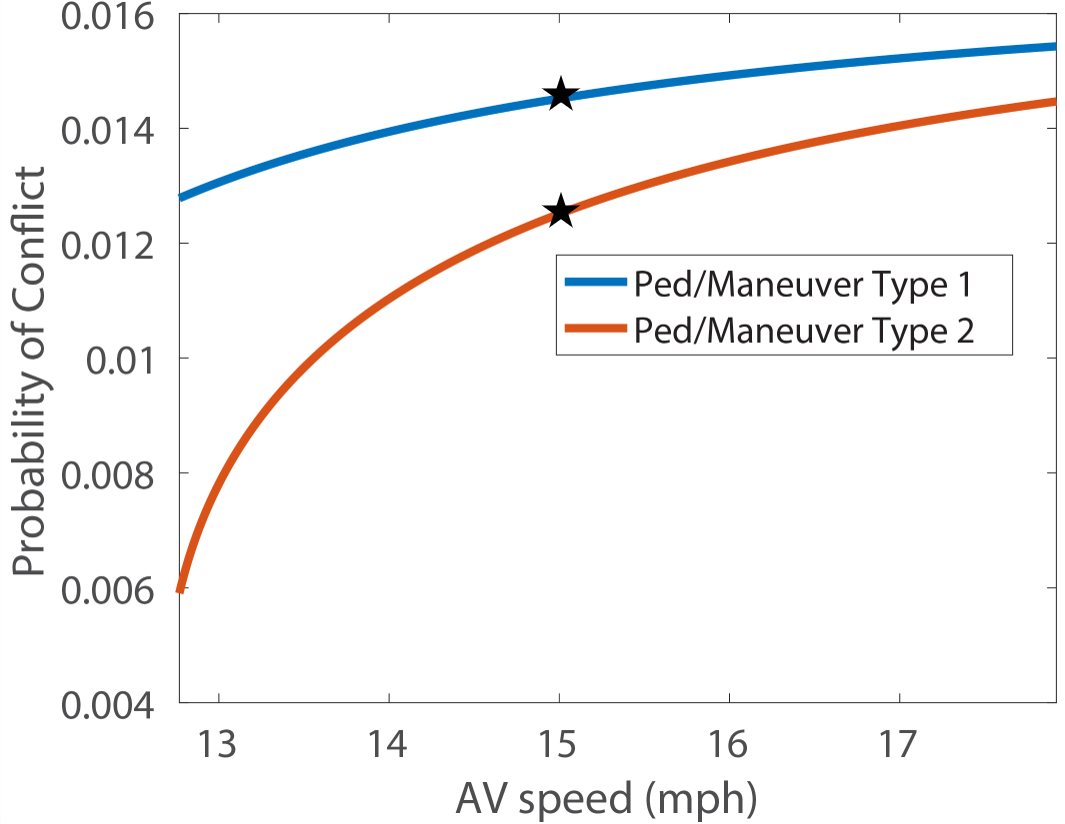}
    \label{fig:ped_maneuver_prob}
  \end{subfigure} \hspace{0.7cm}
  \begin{subfigure}[b]{0.4\linewidth} 
    \includegraphics[width=\linewidth]{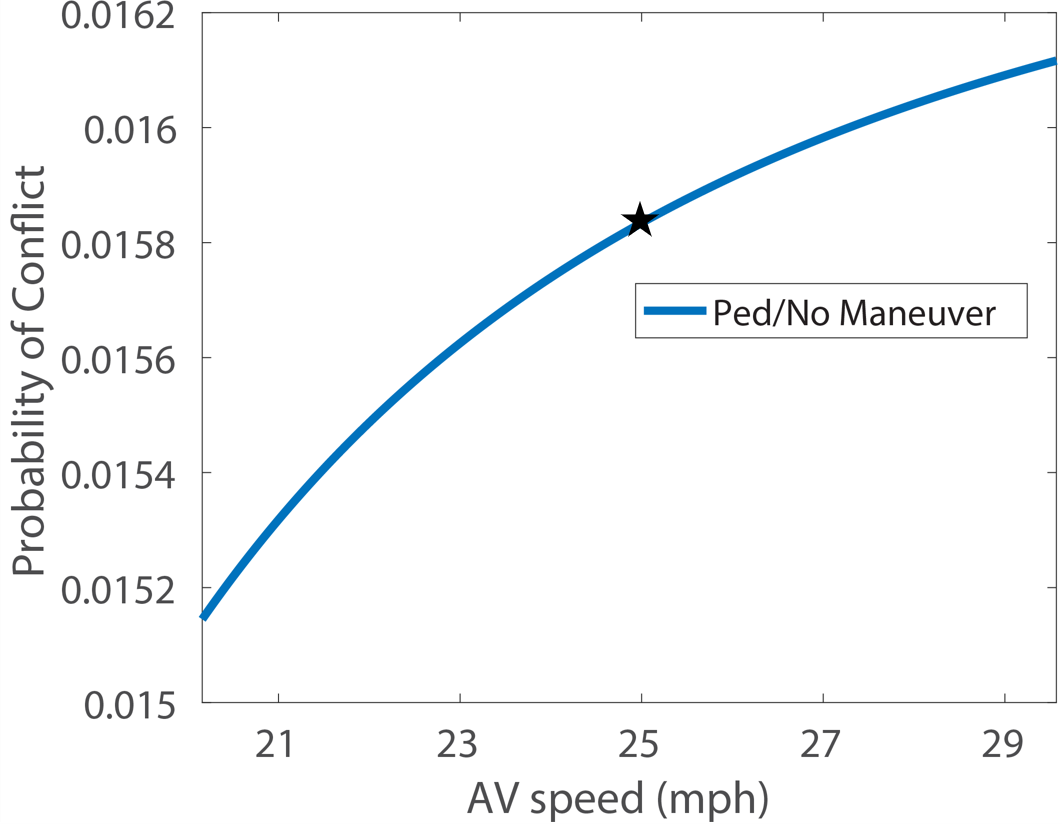}
    \label{fig:ped_no_maneuver}
  \end{subfigure} 
  \caption{Conflict probability as a function of AV's speed for the three occluded pedestrian scenarios. The black star ($\star$) denotes the typical values used in our analysis.}
  \label{fig:ped_conflict_prob}
\end{figure*}

So far, we have calculated conflict probabilities for the three scenarios which are in the range $[0.0125, 0.0158]$. Moreover, Figure \ref{fig:ped_conflict_prob} shows how the conflict probabilities change under varying AV speed $v_{\mathrm{AV}}$. As in \eqref{eq:conflict_to_collision}, this conflict probability can be translated into a collision probability as follows:
\begin{align*}
    P(\text{Collision $|$ Conditions 1 and 2 hold})= \frac{P(\text{Conflict $|$ Conditions 1 and 2 hold})}{\gamma} ,
\end{align*}
where $\gamma$ is the conflict-to-collision ratio for the corresponding pedestrian occlusion scenario. Although we are not aware of any empirical estimates for $\gamma$ specific to scenarios involving pedestrians, it is reasonable to expect that $\gamma$ should be lower than that for the various scenarios considered in \cite{glauz1985expected}. There are two reasons for this: (i) the empirical estimates are not conditioned on occlusions, and  (ii) we account for basic evasive maneuvers such as braking by the AV. Even if we set $\gamma$ as the largest observed value across all scenarios in \cite{glauz1985expected}, we have
\begin{align}
    P(\text{Collision $|$ Conditions 1 and 2 hold}) \approx 2.8 \times 10^{-6},
\end{align}
which is significantly high considering that such occlusions are commonplace in urban driving situations. Due to the reasons mentioned above, we can expect this collision probability to be considerably higher. This strongly suggests that such occlusions are a major cause of crashes involving pedestrians.

\subsection{How can such crashes be prevented?}
We saw that under the occlusion scenarios discussed so far, an AV cannot guarantee safety even if it is willing to accept a crash risk comparable to empirically observed crash rates. The underlying cause of such crashes is the inability to detect vehicles or pedestrians on conflicting paths in time to take evasive action. Such occlusion scenarios are common during peak hours in dense urban environments. As such, waiting for the occlusion to clear before making the maneuver results in a substantial loss in throughput. Thus, mitigating such crashes without sacrificing throughput requires additional communication between vehicles and the infrastructure so that this critical information is relayed to the necessary parties in time. The vehicle-vehicle scenario illustrated in Figure \ref{fig:veh_veh_occlusion} can be prevented by placing a sensor on the through-moving lane at a sufficient distance from the conflict zone depending on the through moving vehicle speeds. For example, if the speed limit on the through moving lane is $30$ mph and  the worst case human reaction time $\rho = 2.5$ s, a sensor placed at 56 m from the conflict zone will provide enough time for the TMV to prevent the impending crash \cite{misener2010cooperative}. In the vehicle-pedestrian scenario illustrated in Figure \ref{fig:ped_scenario2}, an additional sensor detecting pedestrian movement on the crosswalk would be required to prevent the crash.
\section{Traffic Violation}
\begin{figure}[!ht]
\centering
\includegraphics[width=0.8\textwidth]{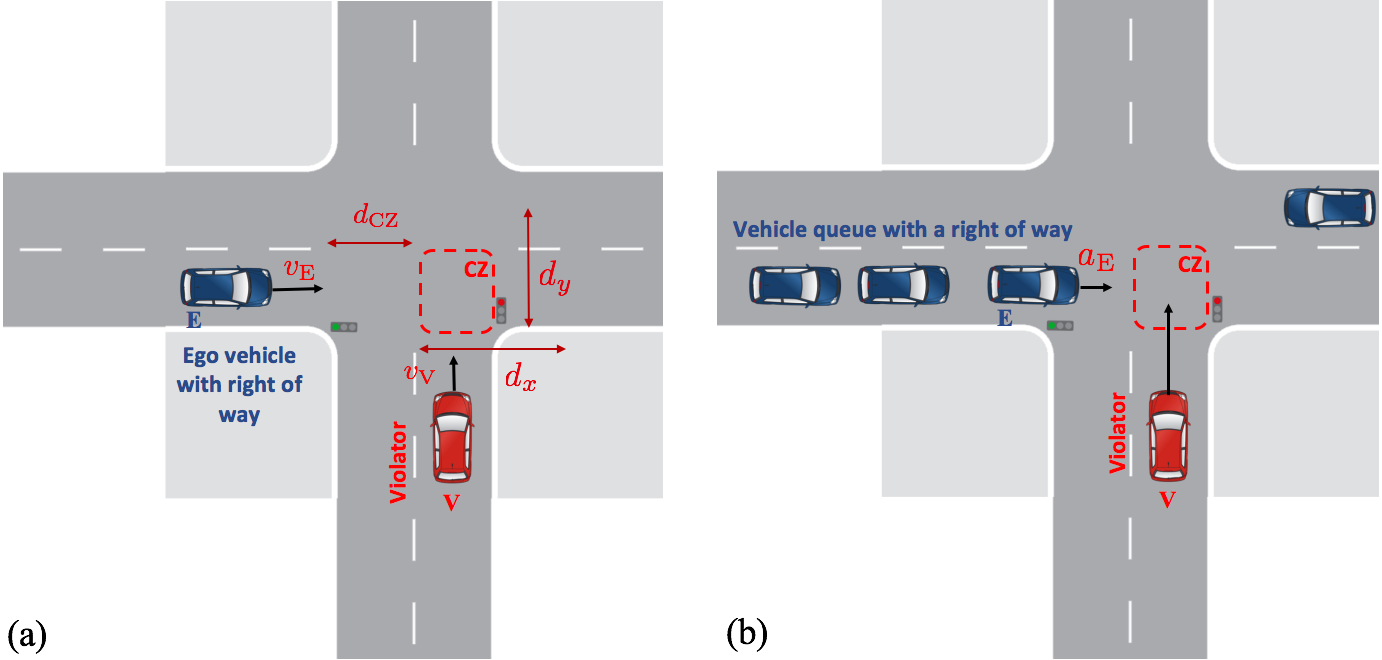}
\caption{Two scenarios of red light violations:
(a) vehicles with a right of way randomly arrive during green time;
(b) vehicles with a right of way wait in queue and slowly start moving as the light changes to green.
}\label{fig:violations}
\end{figure}

Traffic violation is one of the leading causes of crashes on the roads, accounting for 32\% of all crashes \cite{swanson2019statistics}. Red light running is responsible for a large number of crashes each year. Figure~\ref{fig:violations} depicts the setup for a conflict resulting from this violation.
Vehicle V (violator) is going with speed $v_{\mathrm{V}}$ from south to north and runs the red light.
Vehicle E (ego-vehicle) has the right of way traveling from west to east. In the following analysis, we compute the probability of a conflict between
vehicles E and V. Recall that a conflict does not mean crash, but rather a hazardous situation that may lead to a crash,
as discussed in Section~\ref{sec:occlusions}.
We consider the violation scenario from the point of view of the ego-vehicle.
Therefore, we compute the conflict probability under the condition that 
vehicle E is present at the intersection.

\begin{figure}[!ht]
\centering
\includegraphics[width=1.0\textwidth]{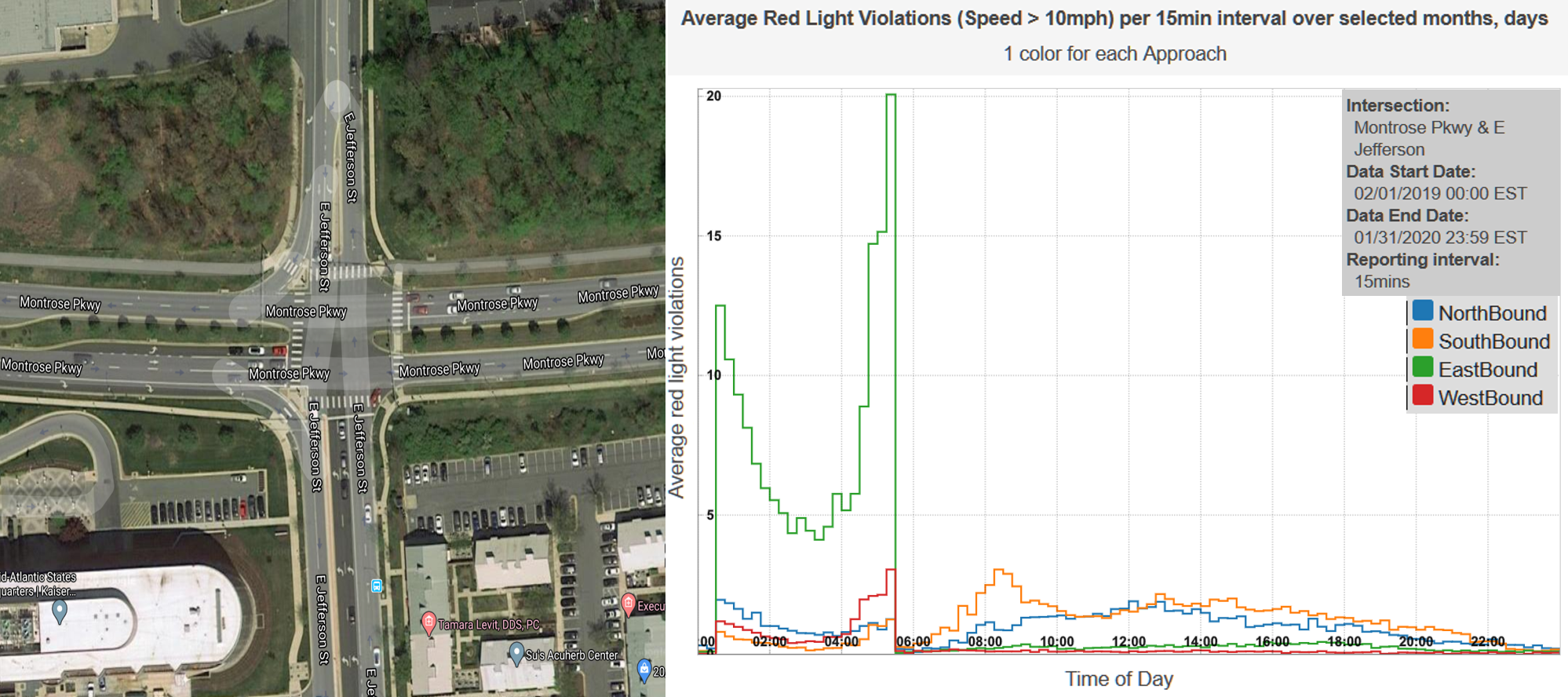}
\caption{Intersection of Montrose Parkway and E. Jefferson Street
in North Bethesda, MD:
Statistics of red light violations collected over one year
from 02/01/2019 through 01/31/2020.}\label{fig:montrose_violations}
\end{figure}

As discussed in~\cite{muralidharan16}, at intersections equipped with stop bar detectors
and a conflict monitoring card that provides programmatic access to a signal phase, it 
is possible to monitor red light violations and collect corresponding statistics.
An example of such data, collected at the intersection of Montrose Parkway and E. Jefferson Street
in North Bethesda, MD, is presented in Figure~\ref{fig:montrose_violations}.\footnote{Note
the spike of violations in the east and less in the west directions between midnight and 6 A.M.
This could be explained by the fact that the southbound direction leads to  Kaiser hospital,
and during the night the south-to-north approach has practically no traffic.
Hence, eastbound and westbound violators feel relatively safe, not expecting danger from that approach
during night hours.
Shortly before 6 A.M. the situation changes sharply, as traffic to the hospital starts to increase.}
Let $\nu_{\mathrm{A}}(t)$ denote the expected number of violations for a given approach A and a given time $t$.
On the south-to-north (northbound -- NB) approach, we have $\nu_{\mathrm{NB}}=0.67$ 
violations between 4:00 and 4:15 A.M.,
and $\nu_{\mathrm{NB}}=1.91$ violations between 12:00 and 12:15 P. M..
These values for the given time intervals
are averaged over a period of one year, from 02/01/2019 through 01/31/2020.

We assume that a red light violation occurs shortly after the signal phase change:
when the green (or yellow) light for vehicle V switches to red.\footnote{Violating late in
the red phase would imply a malicious intent, a rather rare occasion, which we do not consider.}
Suppose, an average signal cycle length is $T_c$, so over a period $\Delta T$ we expect to have
$\Delta T / T_c$ green-to-red switches.
Then, the probability of a violation from approach A during a green-to-red switch is 
\begin{equation}
p_{\mathrm{A}}^{\mathrm{V}}(t) = \frac{T_c\nu_{\mathrm{A}}(t)}{\Delta T}.
\label{eq:violation_prob}
\end{equation}

Thus, given the cycle length $T_c=150$ seconds at our intersection of Montrose Parkway / E. Jefferson Street
and $\Delta T=900$ seconds, we get
$p_{\mathrm{NB}}^{\mathrm{V}}=0.67 / 6=0.11$ between 4:00 and 4:15 A.M. and
$p_{\mathrm{NB}}^{\mathrm{V}}=1.91 / 6=0.32$ between 12:00 and 12:15 P.M.

Let us now discuss the probability of a conflict between vehicles E and V.
Such a conflict, if it occurs, happens in the conflict zone CZ, indicated by the dashed box
in Figure~\ref{fig:violations}.
Let $d_y$ denote the south-to-north size of this conflict zone -- i.e., the distance vehicle V
has to cover being exposed to a conflict with vehicle E.
This distance includes the expected length of vehicle V itself.
In our sample intersection, $d_y=17$ m (includes the length of vehicle, taken to be 4 m).
The time taken by vehicle V to cross the conflict zone is
\begin{equation}
t_{\mathrm{cross}} = t_d + d_y / v_{\mathrm{V}}.
\label{eq:t_cross}
\end{equation}
Here, $t_d$ is the time interval between the signal phase switch that includes red clear time\footnote{Red
clear time is when the signal light for all the intersection movements is red.
It is activated during the signal phase switch and
exists for safety purposes -- to allow everyone who is already inside the intersection to complete their
maneuvers before movements from different approaches are given the right of way.
The typical duration of the red clear period is 2-3 seconds.}
and the instant when vehicle V reaches the conflict zone CZ.
The arrival of vehicle E at the intersection during the violator's crossing can happen in two ways:
\begin{itemize}
\item[(a)] Vehicle E drives through the intersection at the moment of the signal phase switch and,
as its light turns from red to green, at a constant speed without stopping.
This case is depicted in Figure~\ref{fig:violations}(a).
\item[(b)] Vehicle E is already there -- waiting at the red light.
And, as soon as the light turns green, it starts moving.
This case is depicted in Figure~\ref{fig:violations}(b).
\end{itemize}
Since we compute the conflict probability under the assumption that the ego-vehicle is
present in the intersection, for both cases (a) and (b), $p_{\mathrm{EB}}^{\mathrm{E}} = 1$.

It remains to compute the probability of a conflict between vehicles V and E
under the condition that vehicle V runs the red light and vehicle E is also
present at the intersection.
Let $d_{\mathrm{CZ}}$ denote the distance from the stop bar of the west-to-east approach
to the conflict zone CZ,
and $d_x$ -- the size of the conflict zone CZ in the west-to-east direction.
Distance $d_x$ includes an expected length of vehicle E.
In our sample intersection, $d_{\mathrm{CZ}}=16$ m and $d_x=16$ m (accounts for the length of vehicle E, taken to be 4 m).

Denote the red clear time $t_{\mathrm{rc}}$, and set
\[
t^*_d = t_d - t_{\mathrm{rc}}.
\]
Here, we assume $t_{\mathrm{rc}}=3$ seconds.
A conflict exists when both vehicles, V and E, are present in the conflict zone CZ simultaneously.
This happens if the following condition holds:
\begin{itemize}
\item Case (a):
\begin{equation}
\frac{d_{\mathrm{CZ}}}{t^*_d + d_y / v_{\mathrm{V}}} < v_{\mathrm{E}} \leq \frac{d_{\mathrm{CZ}} + d_x}{t^*_d},
\label{eq:conflict_condition_a}
\end{equation}
where $v_{\mathrm{E}}$ is the speed with which vehicle E arrives at and goes through the intersection.
The left part of this inequality states that vehicle E reaches conflict zone CZ before the violator E
manages to cross CZ. 
The right part of this inequality implies that vehicle E would not be able to leave conflict zone CZ
before the violator V reaches this conflict zone.
\item Case (b):
\begin{equation}
2\frac{d_{\mathrm{CZ}}}{(t^*_d + d_y / v_{\mathrm{V}})^{2}} < a_{\mathrm{E}}
\leq 2\frac{d_{\mathrm{CZ}} + d_x}{t_d^{*2}},
\label{eq:conflict_condition_b}
\end{equation}
where $a_E$ is the acceleration with which vehicle E starts moving once its light turns green.
Conditions described by the left and the right sides of this inequality are the same as in~\eqref{eq:conflict_condition_a}.
\end{itemize}
It is possible to estimate the probabilities of conditions~\eqref{eq:conflict_condition_a}
and~\eqref{eq:conflict_condition_b} for different delay periods $t_d$,
if we have some idea about the range of values for $v_{\mathrm{V}}$, $v_{\mathrm{E}}$ and $a_{\mathrm{E}}$.

\begin{figure}[!ht]
\centering
\includegraphics[width=0.5\textwidth]{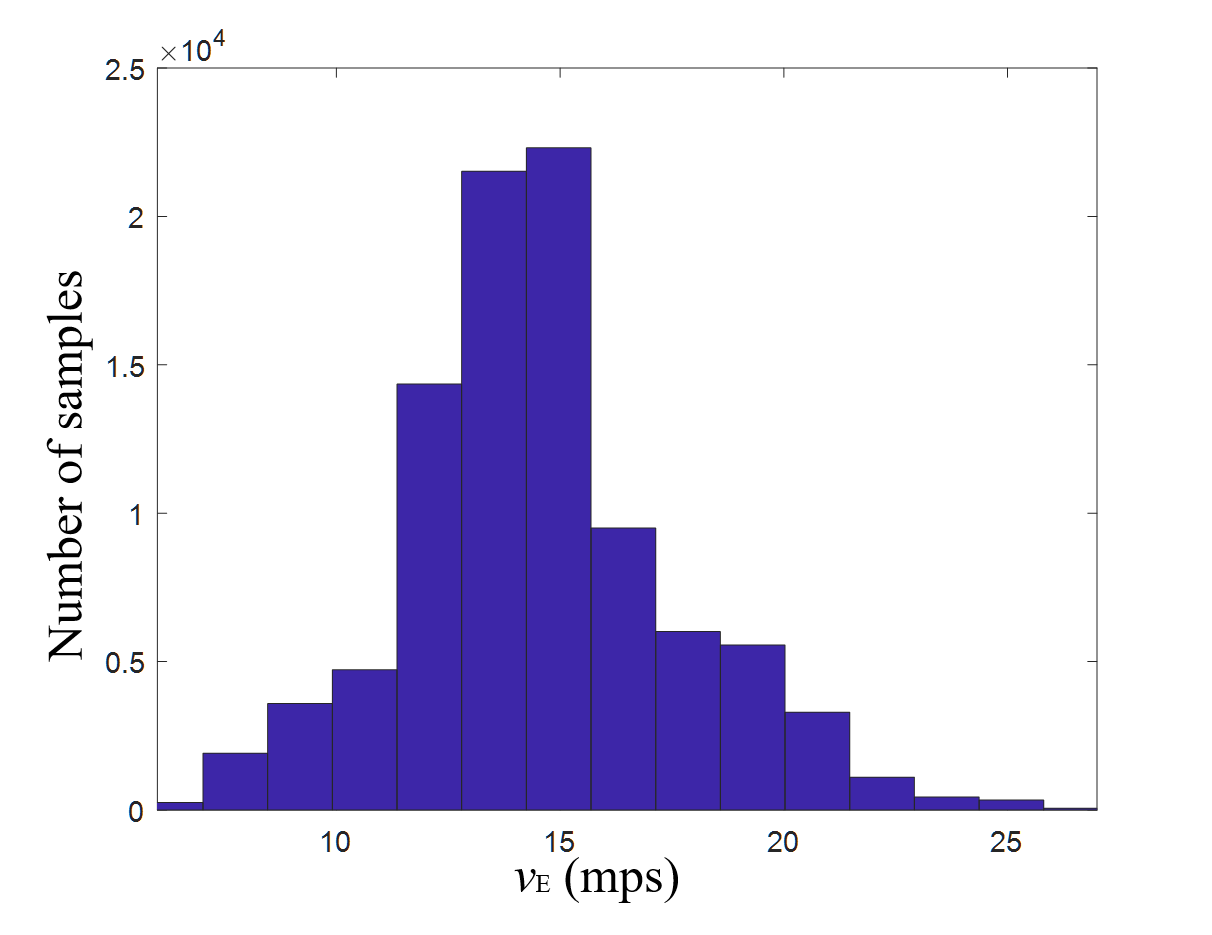}
\caption{Distribution of speeds $v_{\mathrm{E}}$, with which vehicles cross the intersection
of Montrose Parkway and E. Jefferson Street in the west-to-east direction.
The speed values are taken over a random week of 2019.}
\label{fig:e_speed_hist}
\end{figure}

\begin{figure}[!ht]
\centering
\includegraphics[width=1.0\textwidth]{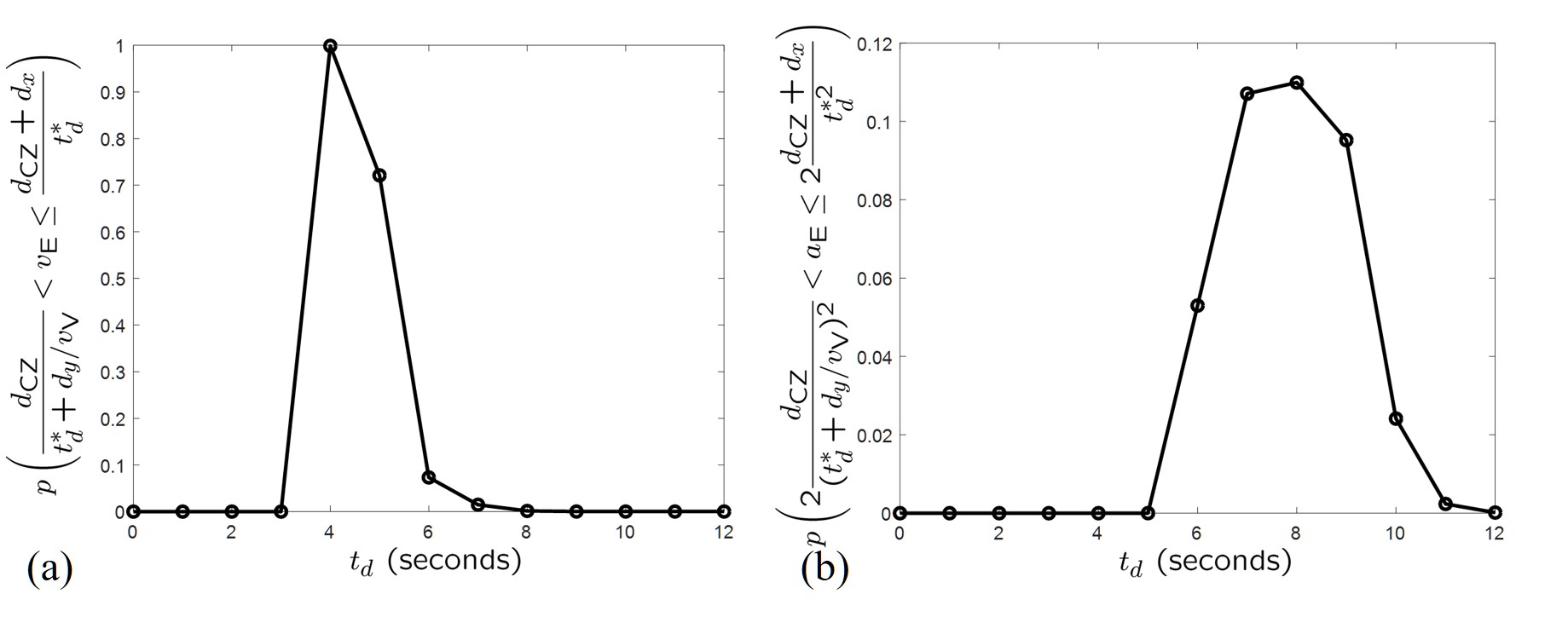}
\caption{Probabilities of a conflict under the condition that both vehicles, V and E,
arrive at the intersection of Montrose Parkway and E. Jefferson Street at the same time -- cases
(a) and (b).}
\label{fig:p_conflict}
\end{figure}

At the intersection of Montrose Parkway and E. Jefferson Street the speed limit in direction
south-to-north is 25 mph (11.18 m/s).
To be more conservative, we will assume the speed of the violator, $v_{\mathrm{V}}=10$ m/s.
The speed of vehicles crossing this intersection in the west-to-east direction measured by
the detectors is presented in the histogram in Figure~\ref{fig:e_speed_hist}.
The typical vehicle acceleration from the intersection stop bar after its light turns green
ranges between 1 and 2 $\mbox{m/s}^2$.
Assuming that $a_{\mathrm{E}}\sim \mathcal{N}(1.5,\,0.25)$, we can estimate the conditional
probabilities of the conflict between vehicles E and V for cases (a) and (b) for different
values of $t_d$ -- see Figure~\ref{fig:p_conflict}.

Finally, we can express the probability of a conflict between vehicles E and V:
\begin{equation}
p_{\mathrm{conflict}} = 
\left\{\begin{array}{ll} p_{\mathrm{NB}}^{\mathrm{V}} 
p\left(
\frac{d_{\mathrm{CZ}}}{t^*_d + d_y / v_{\mathrm{V}}} < v_{\mathrm{E}} \leq \frac{d_{\mathrm{CZ}} + d_x}{t^*_d}
\right), & \mbox{ case (a),} \\
p_{\mathrm{NB}}^{\mathrm{V}}  p\left(
2\frac{d_{\mathrm{CZ}}}{(t^*_d + d_y / v_{\mathrm{V}})^2} < a_{\mathrm{E}}
\leq 2\frac{d_{\mathrm{CZ}} + d_x}{t_d^{*2}}\right), & \mbox{ case (b),}
\end{array}\right.
\label{eq:conflict_probability}
\end{equation}
where $p_{\mathrm{NB}}^{\mathrm{V}}$  is determined from~\eqref{eq:violation_prob}.

\begin{figure}[!ht]
\centering
\includegraphics[width=1.0\textwidth]{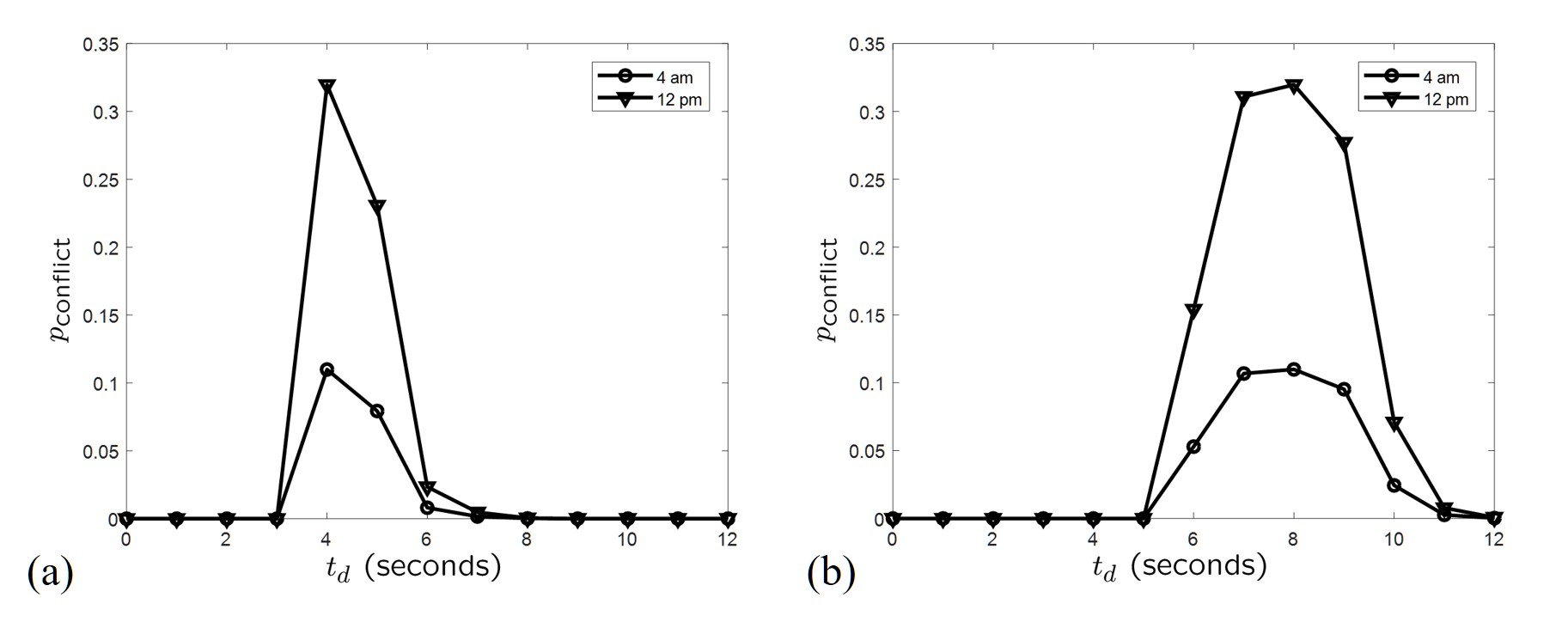}
\caption{Conflict probability computed using data from intersection
of Montrose Parkway and E. Jefferson Street for cases (a) and (b).}
\label{fig:p_conflict_final}
\end{figure}

Given the parameters of our intersection, these probabilities are presented
in Figure~\ref{fig:p_conflict_final} for two time periods: 4:00-4:15 A.M. and 12:00-12:15 P.M..

These conflict probabilities can be translated into collision probabilities using \eqref{eq:conflict_to_collision} with $\gamma = 2040$ estimated for the through-cross traffic scenario in \cite{glauz1985expected}. Based on a similar argument as in Section \ref{sec:occlusions}, this suggests that such crashes will occur with a considerably high probability, which might explain why traffic violations are involved in one-third of all crashes. If we think of the ego-vehicle as an AV attempting to pass the intersection, it is evident that such traffic violations will jeopardize its safety.

The probability of such a conflict could be reduced by I2V technology, provided that vehicle E and the roadside infrastructure were \emph{connected}. The violator V would be detected by combining stop bar sensor activation with a signal phasing.
Then a notification about a violation in a certain direction would be broadcast by the roadside
equipment.
Intersection conflict avoidance (ICA) defined in the SAE J2735 standard~\cite{sae-j2735}
could be used for this purpose.
If vehicle E receives this broadcast before it reaches the conflict zone CZ, it should brake.
If, on the other hand, it is already inside CZ, it should accelerate to clear the conflict zone faster.

\section{Behavior Prediction Uncertainty}
 One of the most challenging aspects of driving on the roads is that one's own safety depends on the behavior of surrounding vehicles. As humans, predicting the actions of vehicles around us is a key component of  driving skill. Attempts are being made to imbue AVs with the same capability by training systems with large amounts of driving data. As there is a vast multitude of driving scenarios and driver behaviors, it is unclear how well such systems generalize to unseen situations. What makes such a prediction task even more challenging is that one's own actions affect surrounding vehicle behavior \cite{sadigh2016planning, schwarting2019social}. For AVs that depend on such behavior prediction modules for planning, inaccuracies in prediction can result in hazardous scenarios. An alternate approach that circumvents the challenge of behavior prediction is to plan based on the worst-case behavior of surrounding vehicles \cite{shalev2017formal}. Although this seems appealing from a safety perspective, it is unclear whether such an approach is feasible in the real world. If not, uncertainty in behavior prediction will indeed result in crashes. In this section, we investigate whether this is the case using empirical data for a common on-road scenario - merging on to a freeway.
 
 Consider an AV attempting to merge from an on-ramp on to a freeway. As in Section \ref{sec:occlusions}, all other vehicles are human driven. The AV decides to merge if the gap between the lead and lag vehicle on the merging lane is large enough. This scenario falls within \emph{Type 16 -- Changing Lanes/Same Direction} in the NHTSA pre-crash scenario typology and accounts for 6.2\% of all crashes. Moreover, this scenario is particularly relevant for AVs as there have been several accounts of them having difficulties merging into traffic \cite{wired_merging}. 
\begin{figure}[ht]
    \centering
    \includegraphics[scale=0.4]{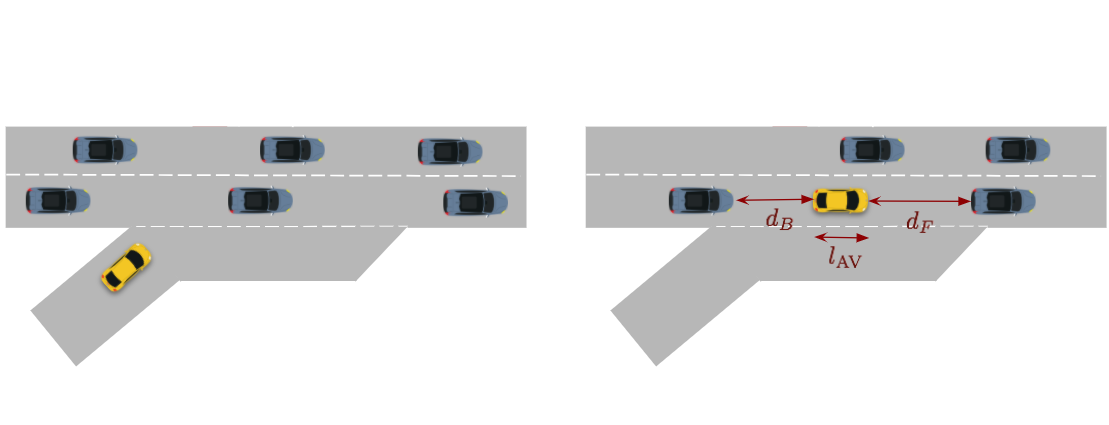}
    \caption{An AV (in yellow) merging from on-ramp on to freeway.}
    \label{fig:merge_onramp}
\end{figure}


We assume that vehicles on the freeway respond to changes in velocity of vehicles in front of them and take evasive action to avoid a crash, albeit with a reaction time. Let $\rho_{\mathrm{AV}}$ and $\rho_{\mathrm{B}}$ denote the reaction times of the AV and lag vehicle respectively. Let $v_{\mathrm{AV}}, v_{\mathrm{F}}$, and $v_{\mathrm{B}}$ denote the speeds of the AV, lead and lag vehicles at the time of the AV's merging decision. Let $d_{\mathrm{F}}$ and $d_{\mathrm{B}}$ denote the gaps between the lead vehicle and AV, and lag vehicle and AV respectively. We assume that all vehicles have the same maximum acceleration and deceleration rates, $a_{\mathrm{acc}}$ and $a_{\mathrm{dec}}$ respectively. In order to guarantee safety, the AV must be able to safely evade any potential collision with the lead or lag vehicle regardless of their actions. This implies that the AV must be safe in the following events: 
\begin{enumerate}
    \item Lead vehicle decelerates to a stop while AV merges,
    \item Lag vehicle accelerates while AV merges.
\end{enumerate}
We first consider the worst case wherein both events occur simultaneously. We then analyze the case in which only the first event occurs -- referred to as the \emph{Single Event} case.

In the worst case, the safe distance for merging $d_{\mathrm{safe}}$ can be decomposed as follows,
\begin{align}
    d_{\mathrm{safe}} = d_{F, \mathrm{safe}} + d_{B, \mathrm{safe}} + l_{\mathrm{AV}},
\end{align}
where $d_{F, \mathrm{safe}}$ and $d_{B, \mathrm{safe}}$ denote the minimum distance required to avoid a collision in events 1 and 2 respectively, and $l_{\mathrm{AV}}$ denotes the length of the AV. In event 1, the lead vehicle brakes to a stop and the AV is able to react only after $\rho_{\mathrm{AV}}$ seconds, before which it maintains its initial velocity $v_{\mathrm{AV}}$. The lead vehicle is initially at a distance $d_{\mathrm{F}}$ from the AV and further traverses a distance $v_{\mathrm{F}}^2/2 a_{\mathrm{dec}}$ before it comes to a stop. The AV travels $v_{\mathrm{AV}} \rho_{\mathrm{AV}}$ until it starts decelerating and then travels a distance $v_{\mathrm{AV}}^2/2 a_{\mathrm{dec}}$ till it stops. Therefore, the AV can safely evade a collision as long as 
\begin{align}
    d_{\mathrm{F}} + \frac{v_{\mathrm{F}}^2}{2 a_{\mathrm{dec}}} \geq  
    v_{\mathrm{AV}} \rho_{\mathrm{AV}} + \frac{v_{\mathrm{AV}}^2}{2 a_{\mathrm{dec}}}.
\end{align}
This gives  
\begin{align}
    d_{F, \mathrm{safe}} = \max \Bigg\{  v_{\mathrm{AV}} \rho_{\mathrm{AV}} + \frac{v_{\mathrm{AV}}^2 - v_{\mathrm{F}}^2}{2 a_{\mathrm{dec}}}, 0 \Bigg\}. \label{eq:df_safe}
\end{align}
In event 2, the lag vehicle accelerates for $\rho_{\mathrm{B}}$ until it realizes that the AV is merging, after which it decelerates to avoid crashing into the AV. The worst case in such a setting is if the AV is forced to brake to a stop because of the lead vehicle doing the same. The AV is initially at a distance $d_{\mathrm{B}}$ from the lag vehicle and travels $v_{\mathrm{AV}}^2/2 a_{\mathrm{dec}}$ further before it comes to a halt. The lag vehicle traverses a distance $v_{\mathrm{B}} \rho_{\mathrm{B}} + \frac{1}{2} a_{\mathrm{acc}} \rho_{\mathrm{B}}^2$ before it starts braking, at which point its speed is $v_{\mathrm{B}} + \rho_{\mathrm{B}} a_{\mathrm{acc}}$. It then travels $(v_{\mathrm{B}} + \rho_{\mathrm{B}} a_{\mathrm{acc}})^2/2 a_{\mathrm{dec}}$ before it comes to a stop. Thus, a collision can be avoided if
\begin{align}
    d_{\mathrm{B}} + \frac{v_{\mathrm{AV}}^2}{2 a_{\mathrm{dec}}} \geq v_{\mathrm{B}} \rho_{\mathrm{B}} + \frac{1}{2} a_{\mathrm{acc}} \rho_{\mathrm{B}}^2 + \frac{(v_{\mathrm{B}} + \rho_{\mathrm{B}} a_{\mathrm{acc}})^2}{2 a_{\mathrm{dec}}}.
\end{align}
Thus, we have 
\begin{align}
    d_{B, \mathrm{safe}} = \max \Bigg\{ 
    v_{\mathrm{B}} \rho_{\mathrm{B}} + \frac{1}{2} a_{\mathrm{acc}} \rho_{\mathrm{B}}^2 + \frac{(v_{\mathrm{B}} + \rho_{\mathrm{B}} a_{\mathrm{acc}})^2 - v_{\mathrm{AV}}^2}{2 a_{\mathrm{dec}}}, 0
    \Bigg\}. \label{eq:db_safe}
\end{align}

\begin{table}[]
\begin{tabular}{|c|c|c|c|c|}
\hline
Time        & \begin{tabular}[c]{@{}c@{}}Merging Lane \\ Speed (mph)\end{tabular} & \begin{tabular}[c]{@{}c@{}}Observed Gap \\ (m)\end{tabular} & \begin{tabular}[c]{@{}c@{}}Worst Case:\\ Safe Gap (m)\end{tabular} & \begin{tabular}[c]{@{}c@{}}Single Event: \\ Safe Gap (m)\end{tabular} \\ \hline
7:50 - 8:05 & 29.73                                                               & 43.19                                                       & 93.79                                                              & 52.30                                                                 \\ \hline
8:05 - 8:20 & 25.43                                                               & 35.63                                                       & 84.22                                                              & 46.36                                                                 \\ \hline
8:20 - 8:35 & 21.65                                                               & 27.44                                                       & 75.90                                                              & 41.22                                                                 \\ \hline
\end{tabular}
\caption{Observed and safe merging gaps for the NGSIM US-101 dataset.}
\label{table:merging_gaps}
\end{table}

\begin{figure*}[h!]
  \centering
  \begin{subfigure}[b]{0.34\linewidth} 
    \includegraphics[width=\linewidth]{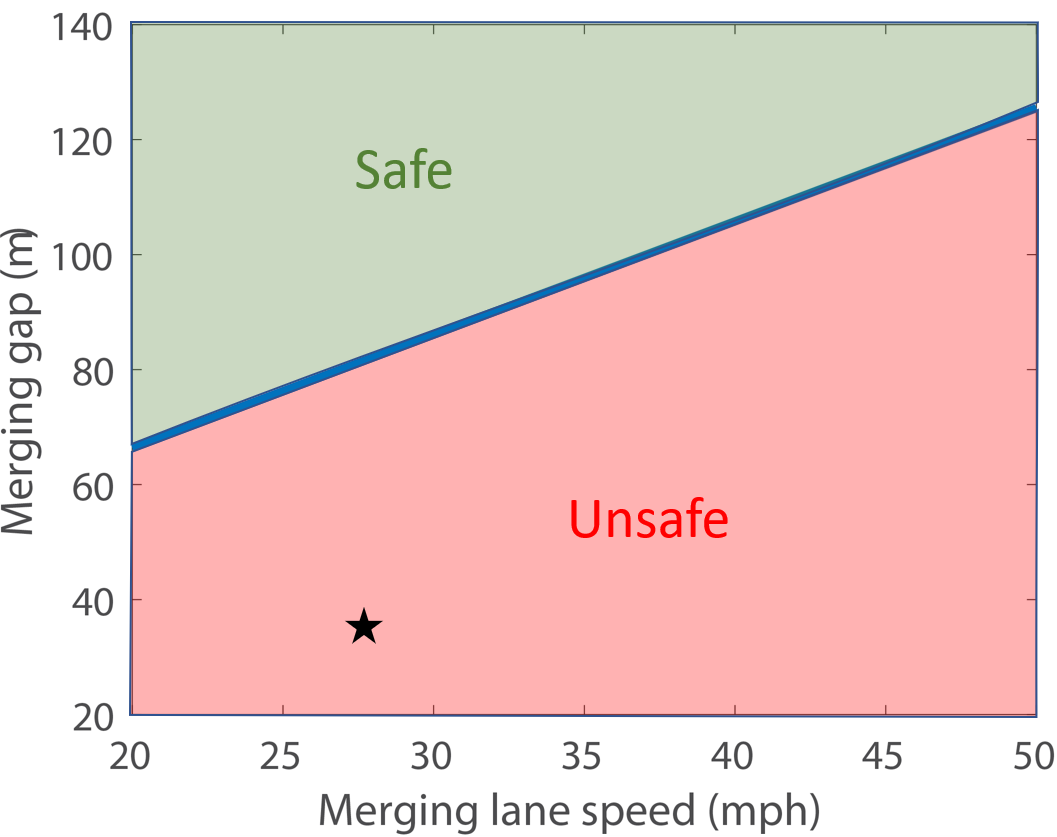}
    \label{fig:merging_safe}
  \end{subfigure} \hspace{1.5cm}
  \begin{subfigure}[b]{0.34\linewidth} 
    \includegraphics[width=\linewidth]{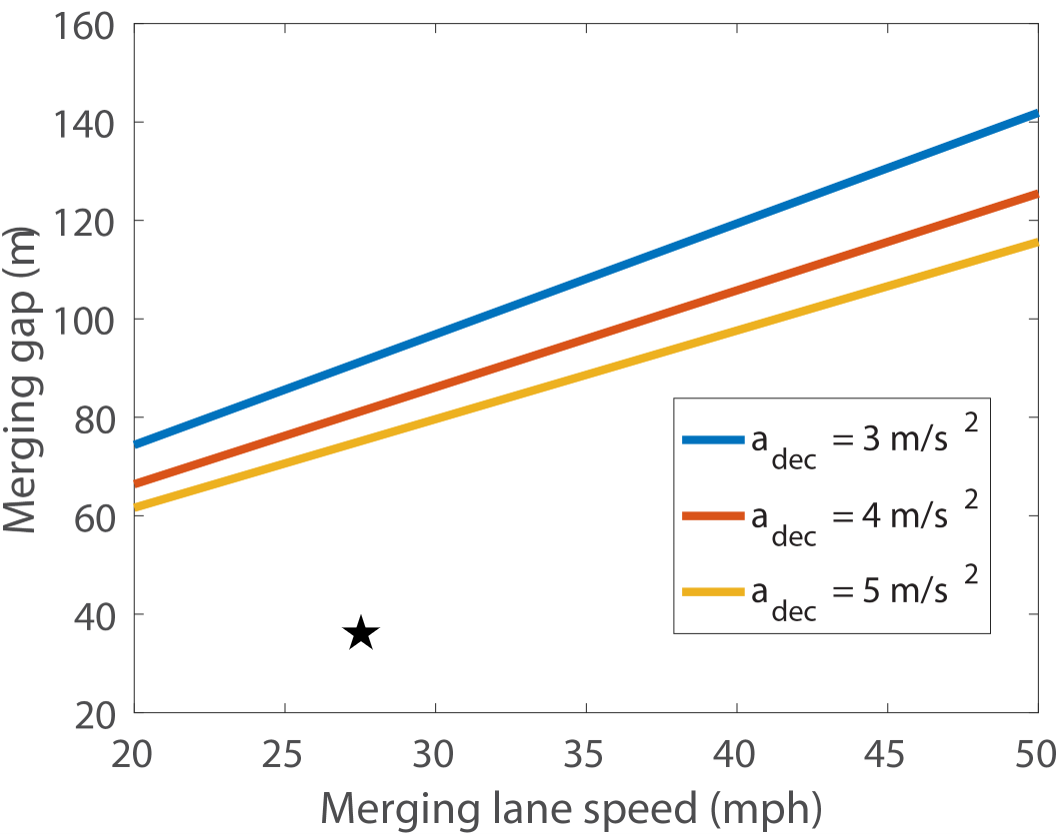}
    \label{merging_sensitivity_dec}
  \end{subfigure} 
  \caption{Safe merging gap as a function of merging lane speed for the time interval 8:05\:AM - 8:20\:AM in the NGSIM US-101 dataset.
  Left: The green (safe) region represents the set of (merging gap, lane speed) pairs for which an AV can merge safely in the worst case, whereas the red (unsafe) region represents unsafe pairs. The black star ($\star$) denotes the observed values of (merging gap, lane speed) for the given time period. Notice that these values are in the unsafe region. \\
  Right: Change in the safe region boundary (as shown on the left) with varying deceleration rates. Note that the observed values from NGSIM remain in the unsafe region for all of the above deceleration rates.}
\end{figure*}

We use the NGSIM US-101 dataset \cite{colyar2007us} to get estimates for vehicle velocities ($v_{\mathrm{AV}}, v_{\mathrm{F}}, v_{\mathrm{B}}$) and observed merging gaps for three 15 minute intervals between 7:50 A.M. and 8:35 A.M. As the AV is accounting for the worst-case, we set the lag vehicle reaction time $\rho_\mathrm{B} = 2.5$ sec according to the American Association of State Highway and Transportation Officials (AASHTO) design specifications based on the $95^{\mathrm{th}}$ percentile of empirical reaction time estimates - see Table 3 in \cite{taoka1989brake}. We set $\rho_{\mathrm{AV}} = 0.83 \ \mathrm{sec}$, which is the mean of empirically observed AV reaction times - see Table 2 in \cite{dixit2016autonomous}. Using the following values for the rest of our problem parameters: $a_{\mathrm{acc}} = 3 \ \mathrm{m/s}^2, \ a_{\mathrm{dec}} = 4 \ \mathrm{m/s}^2$ and $l_{\mathrm{AV}} = 4 \ \mathrm{m}$, we compute the safe merging gap $d_{\mathrm{safe}}$ for each of the time intervals as seen in Table \ref{table:merging_gaps}. Moreover, Figure 12 shows how the safe merging gaps change with varying merging lane speeds and deceleration rates. It can be observed that the observed merging gaps are considerably smaller than the corresponding safe merging gaps. Thus, the worst-case safe planning approach is not feasible in this scenario. Unless the AV predicts the behavior of vehicles in the merging lane accurately, it cannot merge with guaranteed safety. Thus, behavior prediction uncertainty will indeed result in crashes. 

In the above analysis, we considered the worst case of both the lead vehicle braking and lag vehicle accelerating at the same time. We now analyze the \emph{Single Event} case in which only the lead vehicle decelerates to a stop. Observe that the lead gap required to be safe remains the same as in \eqref{eq:df_safe}. However, the lag gap required will be lower in this case as the lag vehicle is not accelerating at the same time. The lag vehicle covers a distance $v_{\mathrm{B}} \rho_{\mathrm{B}}$ before it starts decelerating and further travels $v_{\mathrm{B}}^2/2 a_{\mathrm{dec}}$ before it stops. Thus, the expression for the required lag gap $d_{B, \mathrm{safe}}$ \eqref{eq:db_safe} changes to the following in this case, 
\begin{align}
    d_{B, \mathrm{safe}} = \max \Bigg\{ 
    v_{\mathrm{B}} \rho_{\mathrm{B}} + \frac{v_{\mathrm{B}}^2 - v_{\mathrm{AV}}^2}{2 a_{\mathrm{dec}}}, 0
    \Bigg\}. \label{eq:db_safe_single_event}
\end{align}
As can be seen from Table \ref{table:merging_gaps}, we find that the required gaps even in this case are larger than those commonly observed in the NGSIM dataset.

Deriving estimates for crash probability in this setting is challenging as it requires a model for how vehicles react to surrounding vehicle behavior as well as a probability distribution over possible vehicle behaviors. However, the fact that behavior prediction modules are not always accurate combined with the large number of merging crashes observed each year strongly suggests that such crashes will persist with a significantly high probability.

The fundamental safety challenge in the above merging example is that it requires all the involved vehicles to be in agreement to ensure safety. In the absence of connectivity, predicting vehicle behavior and planning accordingly is the only available recourse. Introducing connectivity between vehicles (V2V) would obviate the need for such behavior prediction and as a result, ensure that the merging maneuver can be executed safely \cite{Hsu1991,luo2016dynamic, yang2004vehicle}.

\section{Conclusion}
Autonomous vehicles (AVs) have the potential to change the transportation landscape and lead us to a safer future.  The thousands of lives that are  lost every year due to impaired, inattentive or reckless driving could be saved if human drivers are replaced with AVs. However, a significant fraction of crashes cannot be explained by these reasons because of a fundamental aspect of driving on the roads: \emph{Our safety depends not only on our own actions but also on the positions and actions of surrounding vehicles -- both of which might be partly unknown to us.} In the absence of connectivity with surrounding vehicles (V2V) or infrastructure (I2V), it is unclear whether AVs can avoid such crashes. Even so, AV companies maintain that they will eventually eliminate all crashes without relying on connectivity. 

One proposed approach is to assume worst case positions and actions of surrounding vehicles and plan  to be safe \cite{shalev2017formal}. While this seems appealing as it circumvents the challenges associated with behavior prediction and does not require connectivity, it turns out that such guaranteed safety comes at a massive cost to traffic efficiency. As we show in our analysis, ensuring guaranteed safety would preclude vehicles from performing basic maneuvers like merging or unprotected left-turns in common traffic scenarios. In the presence of uncertainty about surrounding vehicle positions and behavior, some crashes are indeed unavoidable.

In this paper, we investigate three crash causes to tease out various aspects that make driving on the roads challenging: (a) Occluded Vehicles/Pedestrians (unknown vehicle position), (b) Traffic Violations (rule-following assumption violated), and (c) Behavior Prediction Uncertainty (inability to accurately predict vehicle actions based on past observations). For each of these cases, we show that it is impossible to guarantee worst-case safety without connectivity. We also provide some estimates for the probability of a crash in such scenarios.  

Recognizing that such situations are inevitable while driving on the roads, the objective of an AV should be to manage crash risk while maintaining efficiency. Assessing crash probabilities in commonly occurring risky scenarios is an important step in this process. As we assume perfect sensing and perception capabilities for AVs and idealized scenarios for our analysis, our estimates can be considered as lower bounds for actual crash probabilities. More realistic bounds can be derived by considering imperfections in road-user behavior and AV capabilities, as well as factors such as vehicle failures, and varied road geometry and conditions. Looking through the lens of managing crash risk, AVs can be seen as a collection of Advanced Driver Assistance Systems (ADAS). Thus, empirical estimates of ADAS effectiveness can be used to arrive at better estimates of crash reduction due to AVs \cite{UMTRI_GM}. 

The crashes we have discussed above result from lack of observability of surrounding vehicle positions or incorrect assumptions or predictions about vehicle actions. Communication with surrounding vehicles or infrastructure would ensure that all involved vehicles can detect each other and reach an agreement regarding each others' proposed actions \cite{luo2016dynamic, misener2010cooperative, sae-j2735, yang2004vehicle}. Thus, connectivity can bring about a significant reduction in otherwise unavoidable crashes. However, connectivity comes with its own safety challenges. Relying on information from other vehicles or infrastructure makes one vulnerable to security attacks from malicious road users that can jeopardize safety. Additionally, installing sensors on the roads and ensuring that all vehicles have the required technology to communicate would require a significant amount of time and economic resources.  In order to reap the safety benefits of connectivity, active research is needed to minimize its associated safety risks and economic costs. Our hope for a future with zero crashes depends on it.

\section*{Acknowledgement}
We are grateful to Dr.~Hesham Rakha and Dr.~Kyungwon Kang for providing us with a processed version of the NGSIM US-101 dataset. This research was supported by National Science Foundation EAGER award 1839843, Berkeley Deep Drive, and California Department of Transportation.

\bibliographystyle{plain}
\bibliography{reference}

\newpage
\appendix
\section{Left Turn Occlusion - Analysis of AV's actions}

In this section, we demonstrate why the AV cannot do much to ensure its safety once it decides to make an unprotected left turn. 

According to traffic laws in most jurisdictions, a left turning vehicle must enter the intersection only if it is sure that it can clear it, i.e., it is illegal to enter the intersection with an intention to stop so as to get a better view of through moving vehicles (TMVs). AVs have been observed to be much more conservative compared to human drivers when it comes to obeying traffic rules \cite{av_left_turn_telegraph, av_left_turn_theinfo, verge_krafcik}. We assume that the left-turning AV in our example is programmed to obey traffic rules and hence, will attempt to stop while making the left turn only if it detects a TMV in its field of view. 

 Let $v_{\mathrm{l}}$ denote the AV's left turn speed (starting from rest, it accelerates to $v_{\mathrm{l}}$ and then completes the turn at this speed). To ensure that the AV has the best possible view of TMVs, we assume that there is almost no gap between the AV and the left edge of its lane, whereas all other vehicles are at the center of their lanes. Due to the intersection geometry in this example, the AV traverses a quarter of a circle of radius $R = 9$ m. We use $\theta \in [0, \pi/2]$ to parametrize the point on this circular trajectory at which the TMV comes into the field of view of the AV. At this point, the TMV is at a distance $d_{\mathrm{CZ}} (\theta)$ from the conflict zone. Let us first derive how $d_{\mathrm{CZ}}(\cdot)$ varies as a function of $\theta$. We use the same intersection geometry and vehicle parameters as in Section \ref{sec:occlusions}:
 
 \begin{itemize}
     \item Lane width $l_{\mathrm{w}}- 4$ m,
     \item AV's length $l_{\mathrm{AV}} - 4$ m, 
     \item Maximum acceleration rate $a_{\mathrm{acc}} = 3 \ \mathrm{m/s}^2$, 
     \item Maximum deceleration rate $a_{\mathrm{dec}} = 4 \  \mathrm{m/s}^2$.
 \end{itemize}

Thus, $l(\mathrm{BC}) = 9 \mathrm{cos}\theta - 5$, $l(\mathrm{DE}) = 9 \mathrm{cos}\theta - 1$, and $l(\mathrm{AC}) = 12 - 9 \mathrm{sin}\theta$. As $\triangle$(ABC) and $\triangle$(ADE) are similar, we have
\begin{align}
    \frac{l(\mathrm{BC})}{l(\mathrm{DE})} = \frac{l(\mathrm{AC})}{l(\mathrm{AE})}.
\end{align}
Therefore, $l(\mathrm{AE}) = \frac{(9 \mathrm{cos}\theta - 1)(12 - 9 \mathrm{sin}\theta)}{9 \mathrm{cos}\theta - 5}$. Thus, we have
\begin{align}
    d_{\mathrm{CZ}}(\theta) &= l(\mathrm{CE}), \\
    &= l(\mathrm{AE}) - l(\mathrm{AC}), \\
    &= \frac{4(12 - 9 \mathrm{sin}\theta)}{9 \mathrm{cos}\theta - 5} \label{eq:dcz_theta}.
\end{align}

\begin{figure}
    \centering
    \includegraphics[scale=0.3]{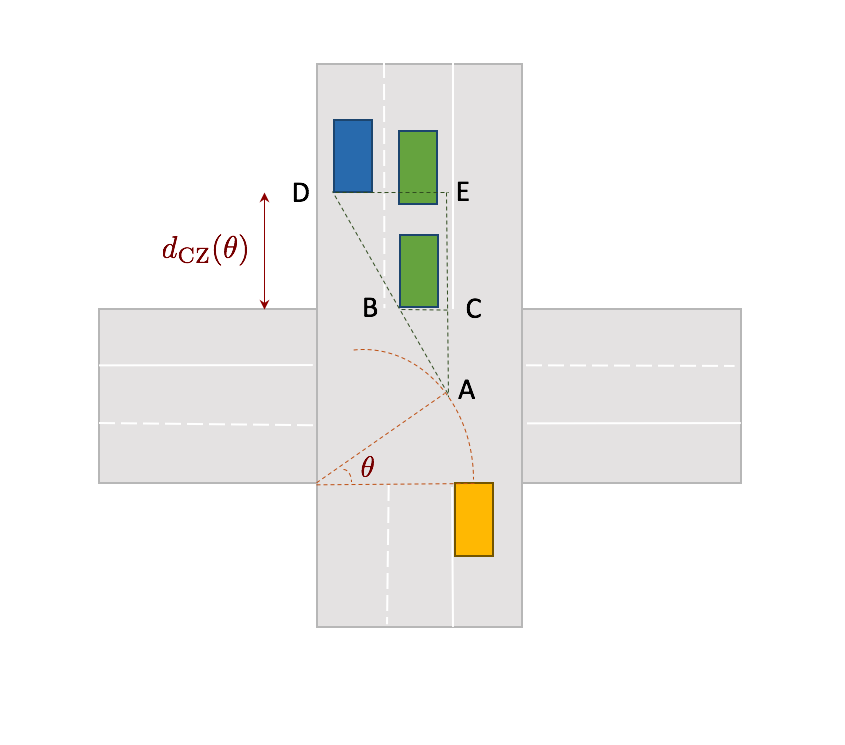}
    \caption{A detailed analysis of the AV's actions to ensure safety during the unprotected left turn.}
    \label{fig:left_turn_detailed}
\end{figure}

Recall that the minimum distance $d^{\mathrm{min}}_{\mathrm{CZ}}$ such that TMVs moving at 25 mph can stop before reaching the conflict zone is equal to 23.45 m. This corresponds to $\theta_{\mathrm{max}} = 0.86$ on the AV's trajectory. If the AV and TMV first come into each other's field of view at any $\theta > \theta_{\mathrm{max}}$, then a crash would be averted as the TMV would have enough time to brake to a stop before the conflict zone. Thus, we can focus on $\theta \leq 0.86$. 

The AV is making the left-turn with velocity $v_{\mathrm{l}} = 4.5$ m/s (10 mph - much lower than the 15 mph speed limit). Let $\rho_{\mathrm{AV}}$ denote the AV's reaction time. Once the TMV comes into its field of view, it takes $\rho_{\mathrm{AV}}$ seconds to take either of the two actions - (i) decelerate to a stop before the conflict zone, and (ii) accelerate to clear the conflict zone before the TMV's arrival. We'll now argue why either of these actions will not ensure the AV's safety. 

Suppose the AV tries to brake to a stop. Let $\theta_{\mathrm{conf}}$ denote the angle corresponding to the AV entering the conflict zone. Then, the AV's distance to the conflict zone when it first sees the TMV is equal to $R (\theta_{\mathrm{conf}} - \theta)$. It travels a distance $v_{\mathrm{l}} \rho_{\mathrm{AV}}$ before it starts decelerating. Thus, it can stop before reaching the conflict zone when
\begin{align}
    \frac{v_{\mathrm{l}}^2}{2 a_{\mathrm{dec}}} \leq R (\theta_{\mathrm{conf}} - \theta) - v_{\mathrm{l}} \rho_{\mathrm{AV}}.
\end{align}
Thus, the AV can be safe in this case only as long as 
\begin{align}
    \theta \leq \theta_{\mathrm{conf}} - \frac{v_{\mathrm{l}}^2}{2 a_{\mathrm{dec}} R} - \frac{v_{\mathrm{l}} \rho_{\mathrm{AV}}}{R} 
\end{align}
On substituting the required parameters, we see that the AV can decelerate to a stop before the conflict zone only when $\theta \leq 0.48$.
On the other hand, if the AV decides to accelerate, it must clear the conflict zone before the TMV's arrival. The TMV is at distance $d_{\mathrm{CZ}} (\theta)$ from the conflict zone when the AV comes into its field of view. It travels a distance $\rho v_{\mathrm{th}}$ before it starts decelerating.
Thus, the time $t_{\mathrm{dec}}$ it takes to reach the conflict zone after it starts decelerating satisfies
\begin{align}
    d_{\mathrm{CZ}} (\theta) = v_{\mathrm{th}} t_{\mathrm{dec}} - \frac{1}{2} a_{\mathrm{dec}} t_{\mathrm{dec}}^2. \label{eq:tmv_dec}
\end{align}
Thus, the total time the AV has in order to clear the conflict zone starting from $\theta$ is given by $t_{\mathrm{TMV}} = \rho + t_{\mathrm{dec}}$. The AV needs to cover the distance $R (\theta_{\mathrm{conf}} - \theta) + l_{\mathrm{w}} + l_{\mathrm{AV}}$ (distance to conflict zone + lane width + AV's length). It covers a distance $v_{\mathrm{l}} \rho_{\mathrm{AV}}$ before it starts accelerating. It can clear the conflict zone within $t_{\mathrm{TMV}}$ seconds only as long as 
\begin{align}
    v_{\mathrm{l}} (t_{\mathrm{TMV}} - \rho_{\mathrm{AV}}) + \frac{1}{2} a_{\mathrm{acc}} (t_{\mathrm{TMV}} - \rho_{\mathrm{AV}})^2 \geq R (\theta_{\mathrm{conf}} - \theta) + l_{\mathrm{w}} + l_{\mathrm{AV}} - v_{\mathrm{l}} \rho_{\mathrm{AV}}. \label{eq:av_acc}
\end{align}
Plugging in the required parameters in \eqref{eq:tmv_dec} and \eqref{eq:av_acc}, we find that the AV will be safe using this approach only when $\theta \geq 0.81$. Combining our findings for both approaches, we can conclude that neither accelerating nor decelerating can ensure that a crash is averted when $\theta \in [0.48, 0.81]$. Thus, the AV cannot be worst-case safe while making a left turn in this scenario even if it attempts an evasive maneuver.

In Section \ref{sec:occlusions}, we did not explicitly account for the AV's evasive maneuvers. This corresponds to a conflict occurring when $\theta \in [0, 0.86]$. Thus, the range of unsafe $\theta$ in the case without evasive maneuvers is $0.86/(0.81-0.48) \approx 2.6$ times that in the case with evasive maneuvers. Recall our result from Section \ref{subsec:veh_veh_occlusion} that the AV needed to wait 443 seconds to be confident enough to make the turn. Assuming the AV tolerates the risk of evasive maneuvers as described above, it would need to wait $443/2.6 \approx 170$ seconds (about 3 min). In other words, it would still have to wait multiple traffic cycles in order to be confident enough to turn. 

\newpage
\section{Detailed Explanation of the Occluded Pedestrians Scenario}

In this section, we provide a more detailed explanation for the analysis presented in Section \ref{subsec:occluded_pedestrians}.

\begin{figure}[ht] 
    \centering
    \includegraphics[scale=0.3]{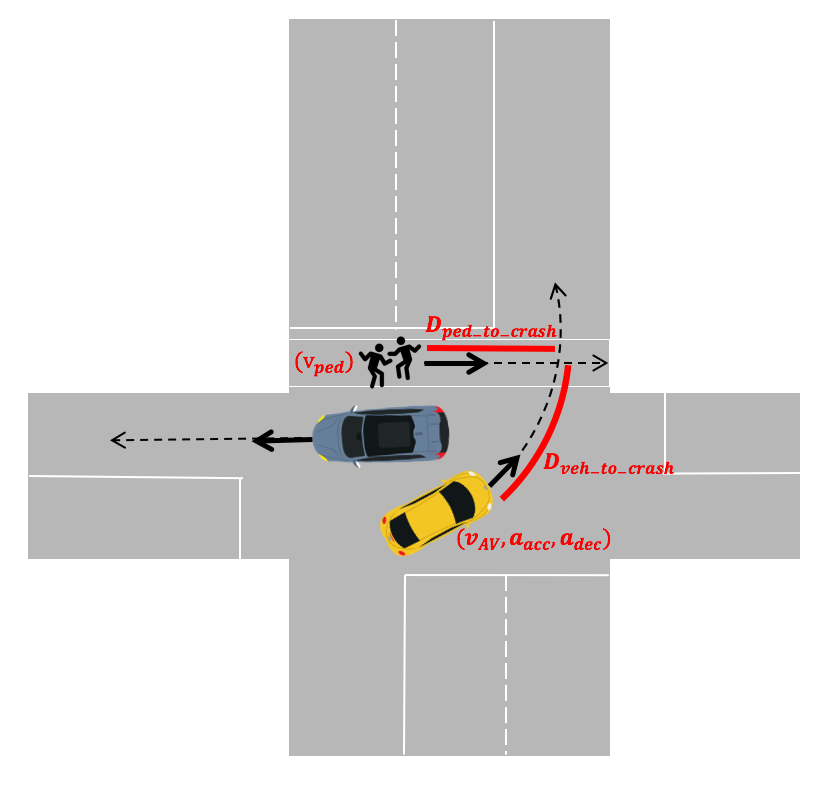}
    \caption{Calculating conflict probability for Pedestrian/Maneuver Type 2.}
    \label{fig:ped_scenario2_app}
\end{figure}

We reproduce here the notation used in Section \ref{subsec:occluded_pedestrians}: 
\begin{itemize}
    \item $\lambda_{\mathrm{ped}}$: arrival rate (ped/s) of pedestrians,
    \item $v_{\mathrm{ped}}$: pedestrian speed (m/s) when crossing on yellow/flashing phase/at the end of pedestrian phase,
    \item $D_{\mathrm{ped\_to\_crash}}$: distance (m) from pedestrian's position at $t=0$ to the center of pedestrian conflict zone,
    \item $v_{\mathrm{AV}}$: speed of the AV (m/s),
    \item $D_{\mathrm{veh\_to\_crash}}$: distance (m) from vehicle's position at $t=0$ to the pedestrian conflict zone,
    \item $a_{\mathrm{acc}}$: maximum AV acceleration rate ($\mathrm{m/s^2}$),
    \item $a_{\mathrm{dec}}$: maximum AV deceleration rate ($\mathrm{m/s^2}$),
    \item $w_{\mathrm{AV}}$: width of AV (m).
\end{itemize}

As discussed in Section \ref{subsec:occluded_pedestrians}, the safety challenge in this scenario is that the AV's field of view of pedestrians on the crosswalk is obstructed by a neighboring vehicle. Once the AV detects the pedestrian, it can take two possible actions to avert a crash:
\begin{enumerate}
    \item Accelerate so that it reaches the conflict zone before the pedestrian enters it,
    \item Decelerate so that it reaches the conflict zone after the pedestrian crosses it.
\end{enumerate} 

We consider $t = 0$ to be the time at which the AV first detects the pedestrian. Let $\delta$ denote the time required by the pedestrian to cross the conflict zone, i.e., $\delta = w_{\mathrm{AV}}/v_{\mathrm{ped}}$. Let $v_{\mathrm{AV}}$ denote the AV's initial speed and let $t_{\mathrm{acc}}$ denote the time taken by the AV to reach the conflict zone while it accelerates. As the distance between the AV and the conflict zone is $D_{\mathrm{veh\_to\_crash}}$ and it is accelerating at rate $a_{\mathrm{acc}}$, we have 
\begin{align}
   D_{\mathrm{veh\_to\_crash}} =  v_{\mathrm{AV}}t_{\mathrm{acc}} + \frac{1}{2}
   {a_{\mathrm{acc}}} t_{\mathrm{acc}} ^2.
\end{align}
Taking the positive root of this quadratic equation gives
\begin{align}
    t_{\mathrm{acc}} = \frac{\sqrt{2a_{\mathrm{acc}}D_{\mathrm{veh\_to\_crash}} + v_{\mathrm{AV}}^2}-v_{\mathrm{AV}}}{a_{\mathrm{acc}}},
    \label{eq:t_acc_app}
\end{align}
At t = $t_{\mathrm{acc}}$, a conflict occurs if the pedestrian is already in the conflict zone. Notice that this corresponds to the pedestrian being at the center of the conflict zone in the interval $\mathcal{T}_{\mathrm{acc}} = [t_{\mathrm{acc}} - \delta/2, t_{\mathrm{acc}} + \delta/2]$. For instance, if the pedestrian is at the center of the conflict zone at $t = t_{\mathrm{acc}} - \delta/2$, then it implies that at $t = t_{\mathrm{acc}}$ (the time at which AV enters the conflict zone), the pedestrian is at the right edge of the conflict zone. Similarly, $t_{\mathrm{acc}} + \delta/2$ corresponds to the pedestrian being at the left edge of the conflict zone. 

On the other hand, let $t_{\mathrm{dec}}$ denote the time taken by the AV to reach the conflict zone while it decelerates at rate $a_{\mathrm{dec}}$. Let $v^{\mathrm{f}}_{\mathrm{AV}}$ denote the the AV's speed when it enters the conflict zone. Then, the following equations hold:
\begin{align}
    v^{\mathrm{f}}_{\mathrm{AV}} &= v_{\mathrm{AV}} - a_{\mathrm{dec}} t_{\mathrm{dec}}, \\
    (v^{\mathrm{f}}_{\mathrm{AV}})^2 &= v_{\mathrm{AV}}^2 - 2a_{\mathrm{dec}}D_{\mathrm{veh\_to\_crash}}.
\end{align}
Solving for $t_{\mathrm{dec}}$, we have
\begin{align}
    t_{\mathrm{dec}} = \frac{v_{\mathrm{AV}}-\sqrt{v_{\mathrm{AV}}^2 -2a_{\mathrm{dec}}D_{\mathrm{veh\_to\_crash}}}}{a_{\mathrm{dec}}},
    \label{t_dec}
\end{align}
and a conflict occurs if the pedestrian arrives at the center of the pedestrian conflict zone in the time interval $\mathcal{T}_{\mathrm{dec}} = [t_{\mathrm{dec}} - \delta/2, t_{\mathrm{dec}} + \delta/2]$. Thus, $\mathcal{T}_{\mathrm{acc}} \cap \mathcal{T}_{\mathrm{dec}}$ represents the time interval in which no matter what action the AV chooses (accelerate or decelerate), a conflict is unavoidable. Since $t_{\mathrm{acc}} < t_{\mathrm{dec}}$, $\mathcal{T}_{\mathrm{acc}} \cap \mathcal{T}_{\mathrm{dec}} = [t_{\mathrm{dec}} - \delta/2, t_{\mathrm{acc}} + \delta/2] $. 

The probability of this event can be interpreted as the probability of arrival of at least one pedestrian in a time interval of length $(t_{\mathrm{acc}} - t_{\mathrm{dec}} + \delta)$. As the pedestrian arrival process is Poisson with rate $\lambda_{\mathrm{ped}}$, we have 
\begin{align}
    P_{\text{scenario2}}(\text{Conflict $|$ Conditions 1 and 2 hold})
    &=1 - e^{-\lambda_{\mathrm{ped}}(t_{\mathrm{acc}} - t_{\mathrm{dec}} + \delta)}.
    \label{eq:p_scenario2}
\end{align}

Recall that $D_{\mathrm{ped\_to\_crash}}$ denotes the distance from the center of the conflict zone at $t = 0$. When the pedestrian is at the center of the conflict zone at $t = t_{\mathrm{dec}} - \delta/2$ (lower limit of $\mathcal{T}_{\mathrm{acc}} \cap \mathcal{T}_{\mathrm{dec}}$), it implies that the pedestrian was at a distance $(t_{\mathrm{dec}}-\delta/2)  v_{\mathrm{ped}}$ at $t = 0$. Similarly, the pedestrian being at the center of the conflict zone at $t = t_{\mathrm{dec}} + \delta/2$ (upper limit of $\mathcal{T}_{\mathrm{acc}} \cap \mathcal{T}_{\mathrm{dec}}$) implies that the pedestrian was at a distance $(t_{\mathrm{dec}} + \delta/2)  v_{\mathrm{ped}}$ at $t = 0$. Thus, we can conclude that 
\begin{align}
    D_{\mathrm{ped\_to\_crash}} \in [(t_{\mathrm{dec}}-\delta/2)  v_{\mathrm{ped}}), \  (t_{\mathrm{acc}}+\delta/2)v_{\mathrm{ped}}]. \label{eq:D_ped_to_crash_app}
\end{align}

The same analysis can be suitably modified for the other two occluded pedestrian scenarios, as suggested in Section \ref{subsec:occluded_pedestrians}.

\end{document}